\documentclass{article} 
\usepackage{iclr2019_conference,times}

\usepackage{amsmath,amsfonts,bm}

\def\eqref#1{equation~\ref{#1}}

\def\1{\bm{1}}

\DeclareMathAlphabet{\mathsfit}{\encodingdefault}{\sfdefault}{m}{sl}
\SetMathAlphabet{\mathsfit}{bold}{\encodingdefault}{\sfdefault}{bx}{n}

\usepackage{hyperref}
\usepackage{url}
\usepackage{graphicx}
\usepackage{subcaption}

\usepackage{algorithm}
\usepackage{algorithmic}

\usepackage{color}
\newcommand{\rtt}{\textcolor[rgb]{0,0,0}}
\DeclareMathOperator*{\argmaxA}{argmax}

\title{Large-Scale Answerer in Questioner's Mind for Visual Dialog Question Generation}

\author{Sang-Woo Lee, Tong Gao, Sohee Yang, Jaejun Yoo, \& Jung-Woo Ha\\
Clova AI Research, NAVER Corp.\\
\texttt{\{sang.woo.lee,tong.gao,sh.yang,jaejun.yoo,jungwoo.ha\}@navercorp.com}
}

\begin{document}

\maketitle

\begin{abstract}
Answerer in Questioner's Mind (AQM) is an information-theoretic framework that has been recently proposed for task-oriented dialog systems. AQM benefits from asking a question that would maximize the information gain when it is asked.
However, due to its intrinsic nature of explicitly calculating the information gain, AQM has a limitation when the solution space is very large. 
To address this, we propose AQM+ that can deal with a large-scale problem and ask a question that is more coherent to the current context of the dialog. 
We evaluate our method on GuessWhich, a challenging task-oriented visual dialog problem, where the number of candidate classes is near 10K. Our experimental results and ablation studies show that AQM+ outperforms the state-of-the-art models by a remarkable margin with a reasonable approximation. In particular, the proposed AQM+ reduces more than 60\% of error as the dialog proceeds, while the comparative algorithms diminish the error by less than 6\%. Based on our results, we argue that AQM+ is a general task-oriented dialog algorithm that can be applied for non-yes-or-no responses. 
\end{abstract} 

\section{Introduction}

Recent advances in deep learning have led an end-to-end neural approach to task-oriented dialog problems that can reduce a laborious labeling task on states and intents \citep{bordes2017}.
Many researchers have applied sequence-to-sequence models \citep{vinyals2015a} that are trained in a supervised learning (SL) and a reinforcement learning (RL) fashion to generate an appropriate sentence for the task. In SL approaches, given the dialog histories so far, the model predicts the distribution of the responses from the task-oriented system \citep{eric2017,de2017,zhao2018}.
However, the SL approach typically requires a lot of training data to deal with unseen scenarios and cover all trajectories of the vast action space of dialog systems \citep{wen2016}.
Furthermore, because the SL-based model does not consider the sequential characteristic of the dialog, the error may propagate over time that causes an inconsistent dialog \citep{li2017,zhao2016}. To address this issue, RL has been applied to the problem \citep{strub2017,das2017b}. By learning the intrinsic planning policy and the reward function, RL approach enables the models to generate a consistent dialog and generalize better on unseen scenarios. However, these methods struggle to find a competent RNN model that uses back-propagation, owing to the complexity of learning a series of sentences \citep{lee2018}.

As an alternative, \cite{lee2018} have recently proposed “Answerer in Questioner’s Mind” (AQM) algorithm that does not depend on a limited capacity of RNN models to cover an entire dialog.
AQM treats the problems as twenty question games and selects the question that gives a maximum information gain. Unlike the other approaches, AQM benefits from explicitly calculating the posterior distribution and finding a solution analytically. The authors showed promising results in the task-oriented dialog problem, such as GuessWhat \citep{de2017}, where a questioner tries to find an object that is in answerer's mind via a series of Yes/No questions. The candidates are confined to the objects that are presented in the given image (less than ten on average). However, this simplified task may not be general enough to practical problems where the number of objects, questions and answers are typically unrestricted. For example, GuessWhich is a generalized version of GuessWhat that has a greater number of class candidates (9,628 images) and a dialog that consists of sentences beyond yes or no \citep{das2017b}.
Because the computational complexity vastly increases to explicitly calculate the information gain over the size of the entire search space, the original AQM algorithm is not scalable to a large scale problem. More specifically, the number of the unit calculation for information gain in GuessWhat is $10$ (number of objects) $\times$ 2 (Yes/No), while that of GuessWhich is $10,000$ (number of images) $\times$ $\infty$ (answer is a sentence) which makes the computation intractable.

One of the interesting ideas \cite{lee2018} suggested is to retrieve an appropriate question from the training set. 
Retrieval-based models, which are basically discriminative models that select a response from a predefined candidate set of system responses, are often used in task-oriented dialog tasks \citep{bordes2017,seo2017a,liu2017}. It is critical not to generate sentences that are ill-structured or irrelevant to the task. However, such a discriminative approach does not fit well with complicated task-oriented visual dialog tasks, because asking an appropriate question considering the visual context is crucial to successfully tackle the problem. It is noticeable that AQM achieved high performance even with a retrieval-based approach in GuessWhat by making the candidate set of questions form the training set. However, \cite{han2017} pointed out that there exist dominant questions in GuessWhat which can be generally applied to all images (contexts), such as ``is it left?’’ or ``is it human?’’. Since GuessWhich is a more complicated task where questions dominant for the game are less likely to exist, it is another reason why the original AQM is difficult to be applied.

To address this, we propose a more generalized version of AQM, dubbed AQM+. Compared to the original AQM, the proposed AQM+ can easily handle the increased number of questions, answers, and candidate classes by employing an approximation based on subset sampling. 
Particularly, unlike AQM, AQM+ generates candidate questions and answers at every turn, and then selects one of them to ask a question.
Because our algorithm considers the previous history of the dialog, AQM+ can generate a more contextual question. To understand the practicality and demonstrate the superior performance of our method, we conduct extensive experiments and quantitative analysis on GuessWhich. Experimental results show that our model could successfully deal with the answers in sentence and significantly decrease 61.5\% of the error while the SL and RL methods decrease less than 6\% of the error. The ablation study shows that our information gain approximation is reasonable. Increasing the number of sampling by eight times brought only a marginal improvement of percentile mean rank (PMR) from 94.63\% to 94.79\%, which indicates that our model can effectively approximate the distribution over the large search space with a small number of sampling. 
Overall, our experimental results provide meaningful insights on how AQM framework can further provide an additional improvement on top of the SL and RL approaches.

Our main contributions are summarized as follows:
\begin{itemize}
    \item We propose AQM+ that extends the AQM framework toward the more general and complicated tasks. AQM+ can handle a more complicated problem where the number of candidate classes is extremely large.
    \item At every turn, AQM+ generates a question considering the context of the previous dialog, which is desirable in practice. In particular, AQM+ generates candidate questions and answers at every turn to ask an appropriate question in the context.
    \item AQM+ outperforms comparative deep learning models by a large margin in Guesswhich, a challenging task-oriented visual dialog task.
\end{itemize}

\section{Related Works}

A task-oriented visual dialog problem has recently been paid attention in the field of computer vision and natural language processing \citep{kim2017}. GuessWhat is one of the famous task-oriented dialog tasks, where the goal is to figure out a target object in the image through a dialog that the answerer has in mind \citep{de2017}. 
However, GuessWhat is relatively an easy task because it only allows the answer form of yes or no. 
The baseline visual question answering (VQA) model achieves 78.5\%.
In the object guessing task (i.e., GuessWhat task itself),
the state-of-the-art averaged accuracy of SL, RL \citep{zhang2018a}, and AQM \citep{lee2018} reached 44.6\% and 60.8\%, and 72.9\% at the 5th round, respectively. Random guessing baseline has an accuracy of 16.0\% \citep{han2017}, thus RL algorithms achieve 53.3\% error decrease, whereas AQM achieves 67.7\%.

GuessWhich is a cooperative two-player game that one player tries to figure out an image out of 9,628 that another has in mind \citep{das2017b}.
GuessWhich uses Visual Dialog dataset \citep{das2017a} which includes human dialogs on MSCOCO images \citep{lin2014} as well as the captions that are generated.
Although GuessWhich is similar to GuessWhat, it is more challenging in every task including asking a question, giving an answer, and guessing the target class. 
For example, unlike GuessWhat that can be answered in yes or no, the answer can be an arbitrary sentence in GuessWhich. 
Therefore, the VQA task in the Visual Dialog dataset is much studied than the GuessWhat dataset \citep{lu2017,seo2017}. 

Similar to GuessWhat, SL and RL approaches have been applied to solve the GuessWhich task and they showed a moderate increase in performance \citep{das2017b,jain2018,zhang2018b}. However, based on the authors' recent Github implementation\footnote{https://github.com/batra-mlp-lab/visdial-rl} of the papers in ICCV \citep{das2017b}, SL and RL methods have shown that only 6\% of error is diminished through the dialog compared to the zeroth turn baselines which only use generated caption.

\section{Algorithm: AQM+}

\subsection{Problem Setting}

\begin{figure}[t] 
\centering
\includegraphics[width=1.00\textwidth]{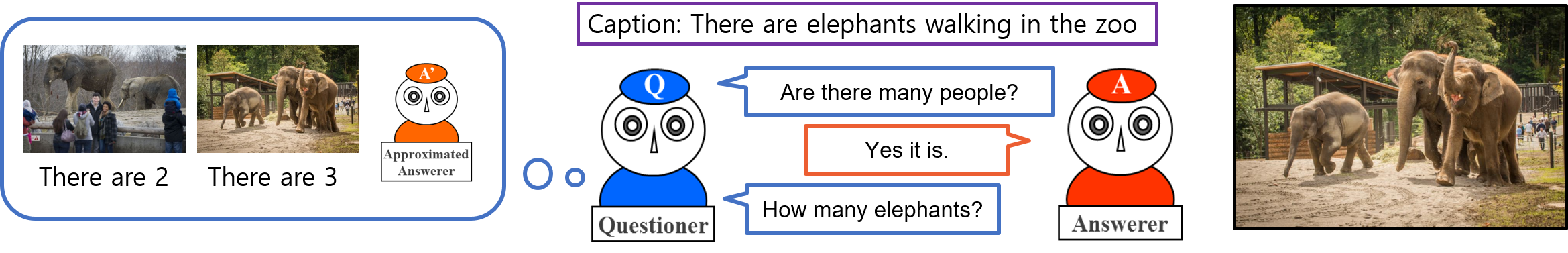}
\caption{Illustration of AQM+ applied for GuessWhich task. The goal of GuessWhich is to figure out a correct answer out of 9,628 test images by asking a sequence of questions. }
\label{fig:iclr19_fig1-1}
\end{figure}

In our experiments, a questioner bot (Qbot) and an answerer bot (Abot) cooperatively communicate to achieve the goal via natural language. Under the AQM framework, at each turn $t$, Qbot generates an appropriate question $q_t$ and guesses the target class $c$ given a previous history of the dialog $h_{t-1}=(q_{1:t-1},a_{1:t-1},h_0)$. Here, $a_t$ is the $t$-th answer and $h_0$ is an initial context that can be obtained before the start of the dialog. 
We refer to the random variables of target class and the $t$-th answer as $C$ and $A_t$, respectively. Note that the $t$-th question is not a random variable in our information gain calculation. To distinguish from the random variables, we use a bold face for a set notation of target class, question, and answers; i.e. $\mathbf{C}, \mathbf{Q}$, and $\mathbf{A}$.

Figure \ref{fig:iclr19_fig1-1} explains the AQM+ algorithm applied to GuessWhich game. In Figure \ref{fig:iclr19_fig1-1}, $c$ is the image with three elephants, $q_1$ is ``Are there many people?'', $a_1$ is ``Yes it is.'', $a_2$ is ``How many elephants?'', and $h_0$ is ``There are elephants walking in the zoo.''
In GuessWhich game, $\mathbf{C}$ is the set of test images whose size is 9,628. The size of $\mathbf{Q}$ and $\mathbf{A}$ is theoretically infinity as questions and answers can be more than one word.

\begin{figure}[t] 
\centering
\includegraphics[width=1.00\textwidth]{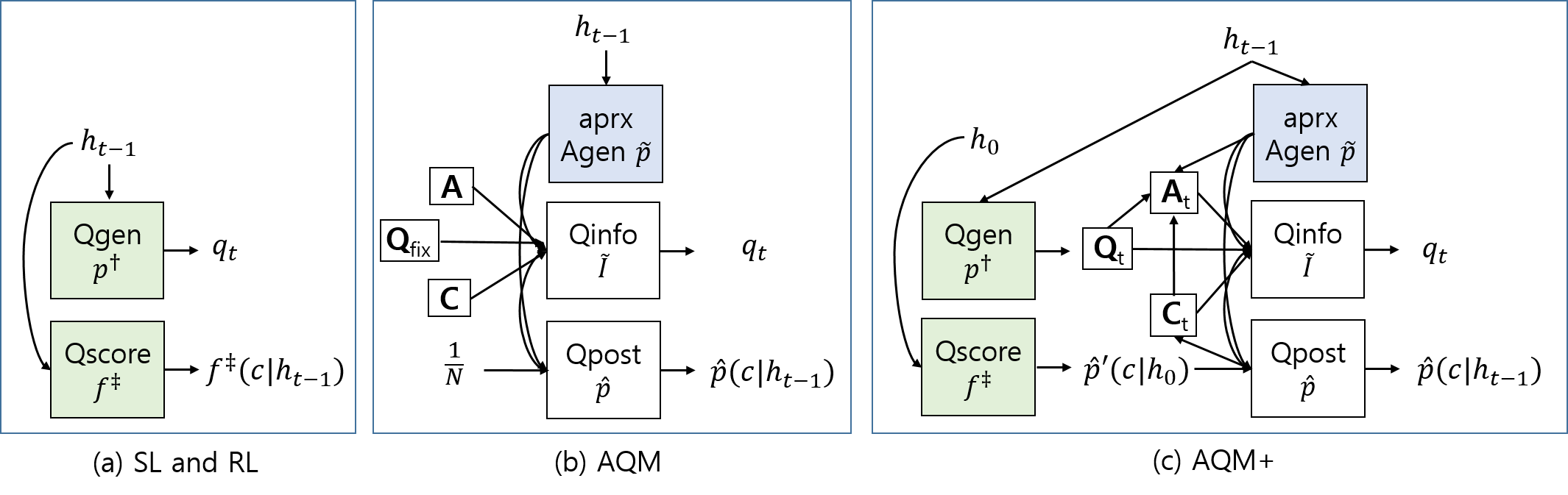}
\caption{\rtt{Architecture of AQM+ and comparative models.} SL and RL have their main neural modules as Qgen $p^\dagger$ and Qscore $f^\ddagger$, while AQM has aprxAgen $\tilde{p}$ used for Qpost \rtt{$\hat{p}$} and Qinfo \rtt{$\tilde{I}$}. AQM+ contains all five modules and uses these to make subsets $\mathbf{Q}_t$, $\mathbf{A}_t$, and $\mathbf{C}_t$, thus achieving approximated estimation on information gain for large-scale inference, along with efficient contextual question generation.}
\label{fig:iclr19_fig1-2}
\end{figure}

\begin{table}[t]
\centering
 \caption{Notation of Qbot's Modules}
    \begin{tabular}{ l  l  l }
     \hline
     \multicolumn{1}{c}{Module} & \multicolumn{1}{c}{Function} & \multicolumn{1}{c}{Explanation} \\ \hline
     Qgen & $p^\dagger(q_t|h_{t-1})$ & a question generating RNN \\ \hline
     Qscore & $f^\ddagger(c|h_t)$ & a score measuring RNN \\ \hline
     aprxAgen & $\tilde{p}(a_t|c,q_t,h_{t-1})$ & an approximated answer generating RNN \\ \hline
     Qinfo & $\tilde{I}[C,A_t;q_t,h_{t-1}]$ & an information gain calculation function by  Equation \ref{eq:eq1} \\ \hline
     Qpost & $\hat{p}(c|h_t)$ & a posterior calculation function by Equation \ref{eq:eq2} \\ \hline \\
  \end{tabular}

 \caption{Notation of Learning Settings}
    \begin{tabular}{ l  l }
     \hline
     \multicolumn{1}{c}{Learning Setting} & \multicolumn{1}{c}{Explanation} \\ \hline
     indA & Like SL, aprxAgen is trained from training data \\ \hline
     depA & Like RL, aprxAgen is trained from the dialog with Abot \\ \hline
     trueA & aprxAgen shares the parameter with Abot \\ \hline
  \end{tabular}
\end{table}

\subsection{Preliminary: SL, RL, and AQM Approaches}
In SL and RL approaches \citep{das2017b,jain2018,zhang2018b}, Qbot consists of two RNN modules. One is ``Qgen", a question generator finding the solution that maximizes its distribution $p^{\dagger}$; i.e. $q^*_t = \argmaxA p^{\dagger}(q_t|h_{t-1})$. 
The other is a ``Qscore'', a class guesser using score function for each class $f^{\ddagger}(c|h_t)$. 
Two RNN modules can either be fully separated two RNNs \citep{strub2017}, or share some recurrent layers but have a different output layer for each \citep{das2017b}.

On the other hand, in the previous AQM approach \citep{lee2018}, these two RNN-based models are substituted to the calculation that explicitly finds an analytic solution. It finds a question that maximizes information gain or mutual information $\tilde{I}$, i.e.  $q^*_t = \argmaxA_{q_t \in \mathbf{Q}_{fix}} \tilde{I}[C,A_t;q_t,h_{t-1}]$, where 
\begin{equation}
\begin{aligned}
& \tilde{I}[C,A_t;q_t,h_{t-1}] = \sum_{c \in \mathbf{C}} \sum_{a_t \in \mathbf{A}} \hat{p}(c|h_{t-1})  \tilde{p}(a_t|c,q_t,h_{t-1}) 
      \ln \frac{\tilde{p}(a_t|c,q_t,h_{t-1})}{\tilde{p}'(a_t|q_t,h_{t-1})}.
\end{aligned}
\label{eq:eq1}
\end{equation}

Here, a posterior function $\hat{p}$ can be calculated with a following equation in a sequential way, where $\hat{p}'$ is a prior function given $h_0$.
\begin{equation}
\hat{p}(c|h_t) \propto \hat{p}'(c|h_0) \prod^t_{j=1} \tilde{p}(a_j|c,q_j,h_{j-1}) = \hat{p}(c|h_{t-1}) \tilde{p}(a_t|c,q_t,h_{t-1})
\label{eq:eq2}
\end{equation}
In AQM, Equation \ref{eq:eq1} and Equation \ref{eq:eq2} can be explicitly calculated from the model. For ease of reference, let us name every component one by one. A module that calculates an information gain $\tilde{I}$ is referred to as ``Qinfo" and a module that finds an approximated answer distribution $\tilde{p}(a_t|c,q_t,h_{t-1})$ is referred to as ``aprxAgen". In AQM, aprxAgen is a model distribution that Qbot has in mind where the target is the true distribution of an answer generator $\bar{p}(a_t|c,q_t,h_{t-1})$, which is referred to as ``Agen''. Finally, ``Qpost" denotes a posterior $\hat{p}$ calculation module for guessing a target class. 

As AQM uses full set of $\mathbf{C}$ and $\mathbf{A}$, the complexity depends on the size of $\mathbf{C}$ and $\mathbf{A}$. For the question selection, AQM uses a predefined set of candidate questions ($\mathbf{Q}_{fix}$), which is not changed for a different turn.

\subsection{AQM+ Algorithm}

In this paper, we propose AQM+ algorithm, which uses sampling-based approximation, for tackling the large-scale task-oriented dialog problem. The core differences of AQM+ from the previous AQM are summarized as follows: 

\begin{itemize}
    \item The candidate question set $\mathbf{Q}_{t,gen}$ is sampled from $p^{\dagger}(q_t|h_{t-1})$ using a beam search at every turn. Previously, \cite{lee2018} used a predefined set of candidate questions $\mathbf{Q}_{fix}$.
    For example, one way to obtain $\mathbf{Q}_{fix}$ is to select questions from the training dataset randomly, called ``randQ".
    \item The answerer model (aprxAgen, $\tilde{p}$) that Qbot has in mind is not a binary classifier (yes/no) but an RNN generator. In addition, aprxAgen does not assume $\tilde{p}(a_t|c,q_t) = \tilde{p}(a_t|c,q_t,h_{t-1})$, which is not even an appropriate assumption when the previous and current questions are sequentially related. For example, $p(a_2=\text{``yes'' }|\text{ }c, q_2=\text{``is left?''}) \neq p(a_2=\text{``yes'' }|\text{ }c, q_2=\text{``is left?''},a_1=\text{``yes''},q_1=\text{``is right?''})$. Regardless of the left term, the probability of the right term is almost zero.
    \item To approximate the information gain of each question, the subsets of $\textbf{A}$ and $\textbf{C}$ are also sampled at every turn. The previous algorithm used full set of $\textbf{A}$ and $\textbf{C}$. We describe an additional explanation on our information gain approximation, infogain\_topk as below.

\end{itemize}

\textbf{Infogain\_topk} The equation for Infogain\_topk is as follows: 
\begin{equation}
\begin{aligned}
& \tilde{I}_{topk}[C,A_t;q_t,h_{t-1}] \\
& = \sum_{a_t \in \mathbf{A}_{t,topk}(q_t)} \sum_{c \in \mathbf{C}_{t,topk}} \hat{p}_{reg}(c|h_{t-1})  \tilde{p}_{reg}(a_t|c,q_t,h_{t-1}) 
      \ln \frac{\tilde{p}_{reg}(a_t|c,q_t,h_{t-1})}{\tilde{p}'_{reg}(a_t|q_t,h_{t-1})},
\end{aligned}
\end{equation}

\noindent
where $\hat{p}_{reg}$ and $\tilde{p}_{reg}$ is a normalized version of $\hat{p}$ over $\mathbf{C}_{t,topk}$ and $\tilde{p}$ over $\mathbf{A}_{t,topk}(q_t)$, respectively. Here, 
$\tilde{p}'_{reg}$ is obtained by using both $\hat{p}_{reg}$ and $\tilde{p}_{reg}$ as follows:

\begin{gather}
\hat{p}_{reg}(c|h_{t-1}) = \frac{\hat{p}(c|h_{t-1})}
{\sum_{c \in \mathbf{C}_{t,topk}} \hat{p}(c|h_{t-1})}\\
\tilde{p}_{reg}(a_t|c,q_t,h_{t-1}) = \frac{\tilde{p}(a_t|c,q_t,h_{t-1})}{\sum_{a_t \in \mathbf{A}_{t,topk}(q_t)} \tilde{p}(a_t|c,q_t,h_{t-1})}\\
\tilde{p}'_{reg}(a_t|q_t,h_{t-1}) = \sum_{c \in \mathbf{C}_{t,topk}} \hat{p}_{reg}(c|h_{t-1}) \cdot \tilde{p}_{reg}(a_t|c,q_t,h_{t-1})
\end{gather}

Each set is constructed by the following procedures.

\begin{itemize}
	\item $\mathbf{C}_{t,topk}$ $\leftarrow$ top-K posterior test images (from Qpost $\hat{p}(c|h_{t-1})$)
    \item $\mathbf{Q}_{t,gen}$ $\leftarrow$ top-K likelihood questions using the beam search (from Qgen $p^\dagger(q_t|h_{t-1})$)
    \item $\mathbf{A}_{t,topk}(q_t)$ $\leftarrow$ top-1 generated answers from aprxAgen for each question $q_t$ and each class in $\mathbf{C}_{t,topk}$ (from aprxAgen $\tilde{p}(a_t|c,q_t,h_{t-1})$)
\end{itemize}

Top-K samples may lead our approximation to be biased toward plausible (high-probability) candidate classes and plausible candidate answers. However, we chose to use top-K samples because our main goal is to reduce the entropy over plausible candidate classes and answers, not over the whole candidate classes and answers.

In general, the AQM+ algorithm can deal with various problems where $|\mathbf{C}_{t,topk}|$, $|\mathbf{Q}_{t,gen}|$, and $|\mathbf{A}_{t,topk}(q_t)|$ are all different. Here, $|\cdot|$ denotes the cardinality of a set. We can vary the size of each set and control the complexity of the AQM+ algorithm. In our experiments, however, we mainly considered the problem when  $|\mathbf{C}_{t,topk}| = |\mathbf{Q}_{t,gen}| = |\mathbf{A}_{t,topk}(q_t)|$. 
More specifically, $|\mathbf{C}_{t,topk}|$ is equal to $|\mathbf{A}_{t,topk}(q_t)|$ because our model finds a single best answer $a_t$ given a pair $(q_t,c)$ that maximizes $\tilde{p}(a_t|c,q_t,h_{t-1})$. 
Therefore, $|\mathbf{A}_{t,topk}|=|\mathbf{Q}_{t,gen}| \cdot |\mathbf{C}_{t,topk}|$ per information gain calculation where $\mathbf{A}_{t,topk}=\{\mathbf{A}_{t,topk}(q_t)|q_t\in \mathbf{Q}_{t,gen}\}$. 
For the detailed explanation, see Algorithm \ref{alg:alg1} in Appendix A.

We also explain the extended sampling method on candidate answers for cases where $\mathbf{A} \neq \mathbf{C}$ is required. In the extended method, aprxAgen first generates top-m answers for each candidate question and each candidate class, where $m$ is the smallest integer satisfying $|A| \leq |C| \cdot m$. After that, the candidate answers are randomly removed, leaving only $|A|$ answers.

\subsection{Learning}

In all SL, RL, and AQM frameworks, Qbot needs to be trained to approximate the answer-generating probability \rtt{distribution} of Abot. 
In AQM approach, aprxAgen does not share the parameters with Agen, and therefore also needs to be trained to approximate Agen. AQM can train aprxAgen by the learning strategy of the SL or RL approach. 
We explain two learning strategies of AQM framework below: indA and depA.
In SL approach, Qgen and Qscore are trained from the training data, which have the same or similar distribution to that of the training data used in training Abot.
Likewise, in indA setting of AQM approach, aprxAgen is trained from the training data.
In RL approach, Qbot uses dialogs made by the conversation of Qbot and Abot and the result of the game as the objective function (i.e. reward). 
Likewise, in depA setting of AQM approach, aprxAgen is trained from the questions in the training data and following answers obtained in the conversation between Qbot and Abot.
We also use the term trueA, referring to the setting where aprxAgen is the same as Agen, i.e. they share the same parameters.
Both the previous AQM algorithm and the proposed AQM+ algorithm use these learning strategies.

\section{Experiments}

\subsection{Experimental Setting}

\textbf{GuessWhich Task} 
GuessWhich is a two player game played by Qbot and Abot. The goal of GuessWhich is to figure out a correct answer out of 9,628 test images by asking a sequence of questions.  
Abot can see the randomly assigned target image, which is unknown to Qbot.
Qbot only observes a caption of the image generated from Neuraltalk2 \citep{vinyals2015a}.
To achieve the goal, Qbot asks a series of questions, to which Abot responds with a sentence.

\textbf{Comparative Models} 
We compare AQM+ with three comparative models, SL-Q, RL-Q, and RL-QA \citep{das2017b}.
In SL-Q, Qbot and Abot are trained separately from the training data.
In RL-Q, Qbot is initialized by the Qbot trained by SL-Q and then is fine-tuned by RL. Abot is the same as the Abot trained by SL-Q, and is not fine-tuned further. In the original paper \citep{das2017b}, it was referred to as Frozen-A.
By the way, in an RL-QA setting, not only Qbot but also Abot is concurrently trained with Qbot. In the original paper, it was referred to as RL-full-QAf.
We also compare our AQM+ with ``Guesser'' algorithm. Guesser asks a question generated from SL-Q algorithm and calculates posterior by Qpost of AQM+.

\textbf{Non-delta vs. Delta Hyperparameter Setting}
The important issue in our GuessWhich experiment is delta setting.
In the paper of \cite{das2017b}, SL-Q, RL-Q, and RL-QA algorithms achieve moderate increases of the performance. In SL-Q, 88.5\% of percentile mean rank (PMR) is improved to 90.9\%.
In RL-QA, 90.6\% of PMR is improved to 93.3\%. 
Here, 93.3\% of PMR at the zeroth turn means that the model can predict the correct image to be more likely than the other 8,983 images out of 9,628 candidates after exploiting the caption information solely.
However, \cite{das2017b} found that another hyperparameter setting, delta, makes much progress on their algorithm. Delta setting refers to different weights on loss and learning decay rate. Based on the authors' recent report on Github, SL-Q and RL-QA methods have shown that less than 6\% of error is diminished through the dialog compared to the zeroth turn baseline which only uses generated caption.
The PMR of the target (class) image which only uses the caption is around 95.5, but the dialog does not improve the PMR to more than 95.8.
We use both non-delta setting (the setting in the original paper) and delta setting (the setting in Github) to test the performance of AQM+.

\textbf{Other Experimental Setting}
As shown in Figure \ref{fig:iclr19_fig1-2}, our model uses five modules, Qgen, Qscore, aprxAgen, Qinfo, and Qpost. 
We use the same Qgen and Qscore modules as the comparative SL-Q model.
In Visual Dialog, Qgen and Qscore share one RNN structure and have different output layers for each.
The prior function is obtained from $\hat{p}'(c|h_0) \propto$ $exp(\lambda \cdot f^\ddagger(c|h_0))$ using Qscore, where $\lambda$ is a balancing hyperparameter between prior and likelihood.
We set $|\mathbf{C}_{t,topk}| = |\mathbf{Q}_{t,gen}| = |\mathbf{A}_{t,topk}(q_t)|$ = 20.
The epoch for SL-Q is 60. The epoch for RL-Q and RL-QA is 20 for non-delta, and 15 for delta, respectively. \rtt{Our code is modified from the code of \cite{modhe2018visdialrlpytorch}, and we make our code publically available\footnote{https://github.com/naver/aqm-plus}.}
\rtt{All experiments are implemented and fine-tuned with NAVER Smart Machine Learening (NSML) platform~\citep{sung2017nsml, kim2018nsml}.}

\begin{table}[t]
\centering
 \caption{Test percentile mean rank (PMR) in 10th round. Caption refers the 0th round PMR of SL-Q. The results of comparative deep models in the non-delta setting is from the paper of \cite{das2017b}.}
    \begin{tabular}{ l  c c c c c c}
     \hline
     & Caption & SL-Q & RL-QA & AQM+ w/ indA & AQM+ w/ depA & AQM+ w/ trueA \\ \hline
    non-delta & 88.5 & 90.9 & 93.3 & 94.64 & 97.45 & 99.87  \\ \hline
    delta & 95.45 & 95.72 & 95.69 & 97.17 & 98.25 & 99.22  \\ \hline
  \end{tabular}
  
  \label{table:table2}

\end{table}

\begin{figure}[t] 
\centering
\begin{subfigure}[b]{0.48\textwidth}
\caption{Non-delta Hyperparameter Setting}
\includegraphics[width=\textwidth]{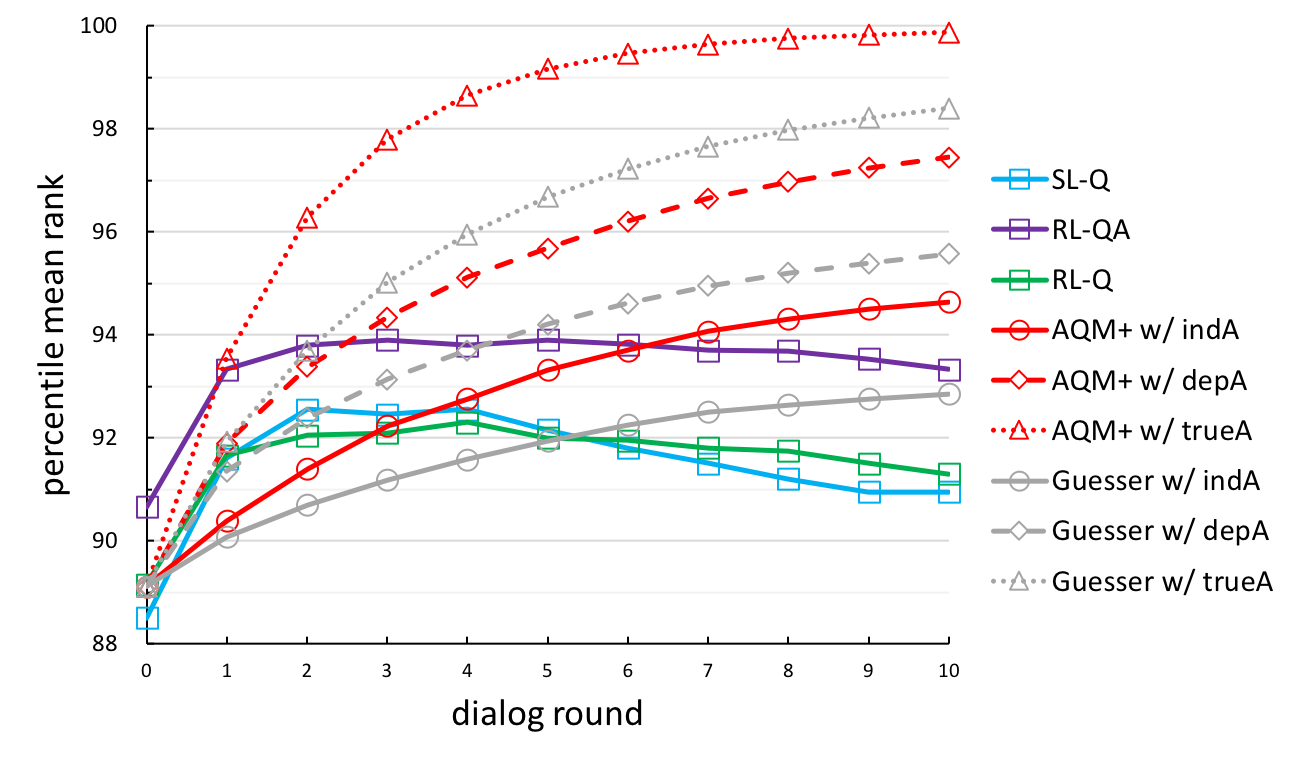}
\end{subfigure}
\begin{subfigure}[b]{0.48\textwidth}
\caption{Delta Hyperparameter Setting}
\includegraphics[width=\textwidth]{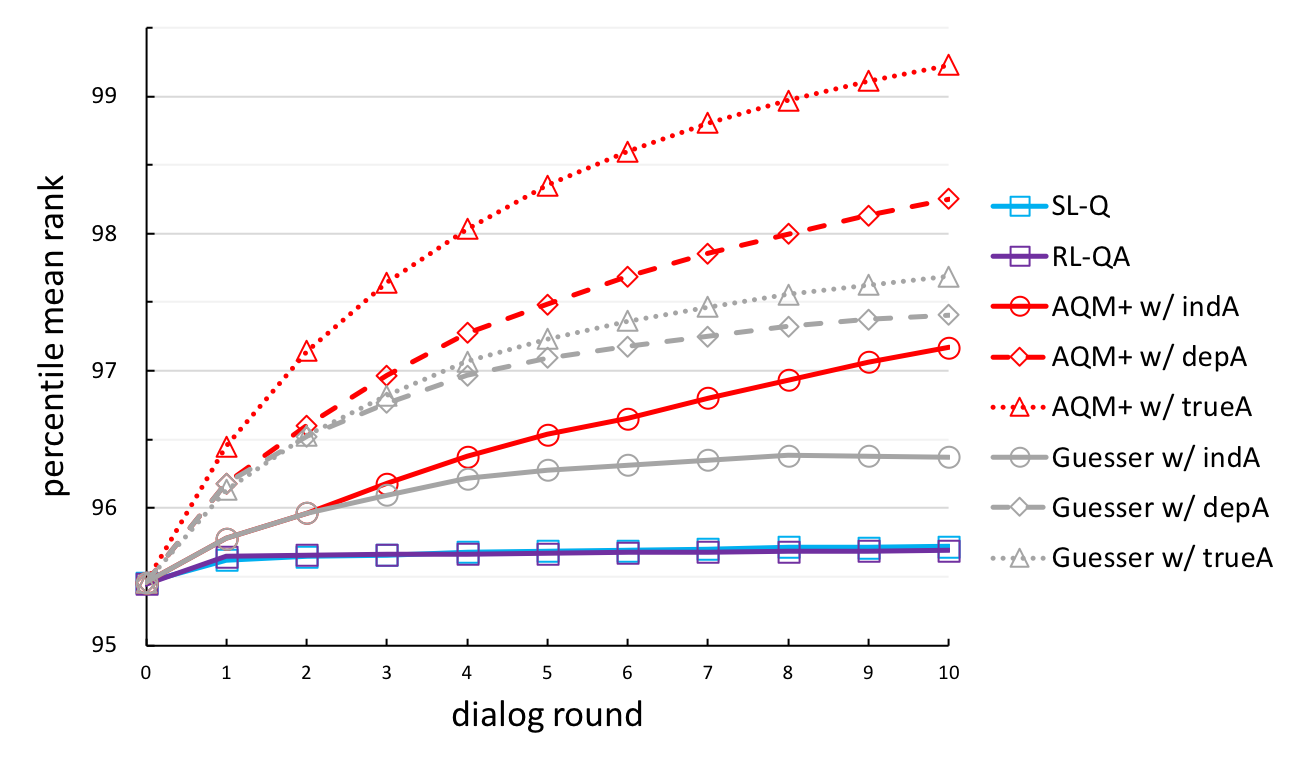}
\end{subfigure}
\caption{Test percentile mean ranks on GuessWhich experiments.}
\label{fig:iclr19_fig2}
\end{figure}

\subsection{Comparative Results}

Figure \ref{fig:iclr19_fig2} shows the PMR of the target image for our AQM+ and comparative models across the rounds.
Figure \ref{fig:iclr19_fig2}a corresponds to the non-delta setting in the original paper \citep{das2017b} and Figure \ref{fig:iclr19_fig2}b corresponds to the delta setting proposed in the Github code.

We see that SL-Q and RL-QA do not significantly improve the performance after a few rounds, especially for the delta setting.
In delta setting, SL-Q increases their performance from 95.45\% to 95.72\% at 10th round, and RL-QA increases their performance from 95.44\% to 95.69\%. It means that error drop of SL-Q and RL-QA algorithms is 5.74\% and 5.33\%, respectively.
On the other hand, AQM-indA increases its PMR from 95.45\% to 96.53\% at the fifth round and reaches 97.17\% at the 10th round. 
Likewise, AQM-depA increases its PMR from 95.45\% to 97.48\% at the fifth round and reach 98.25\% at the 10th round, decreasing 61.5\% of error.
Note that Guesser w/ indA achieves 96.37\% at the 10th round, outperforming SL-Q by a significant margin. It shows that not only the question generation but also the guessing mechanism affects the performance degeneration of SL and RL algorithms.

\subsection{Ablation Study}

\begin{figure}[t] 
\centering
\begin{subfigure}[b]{0.48\textwidth}
\caption{No Caption Experiment (indA, Non-delta)}
\includegraphics[width=\textwidth]{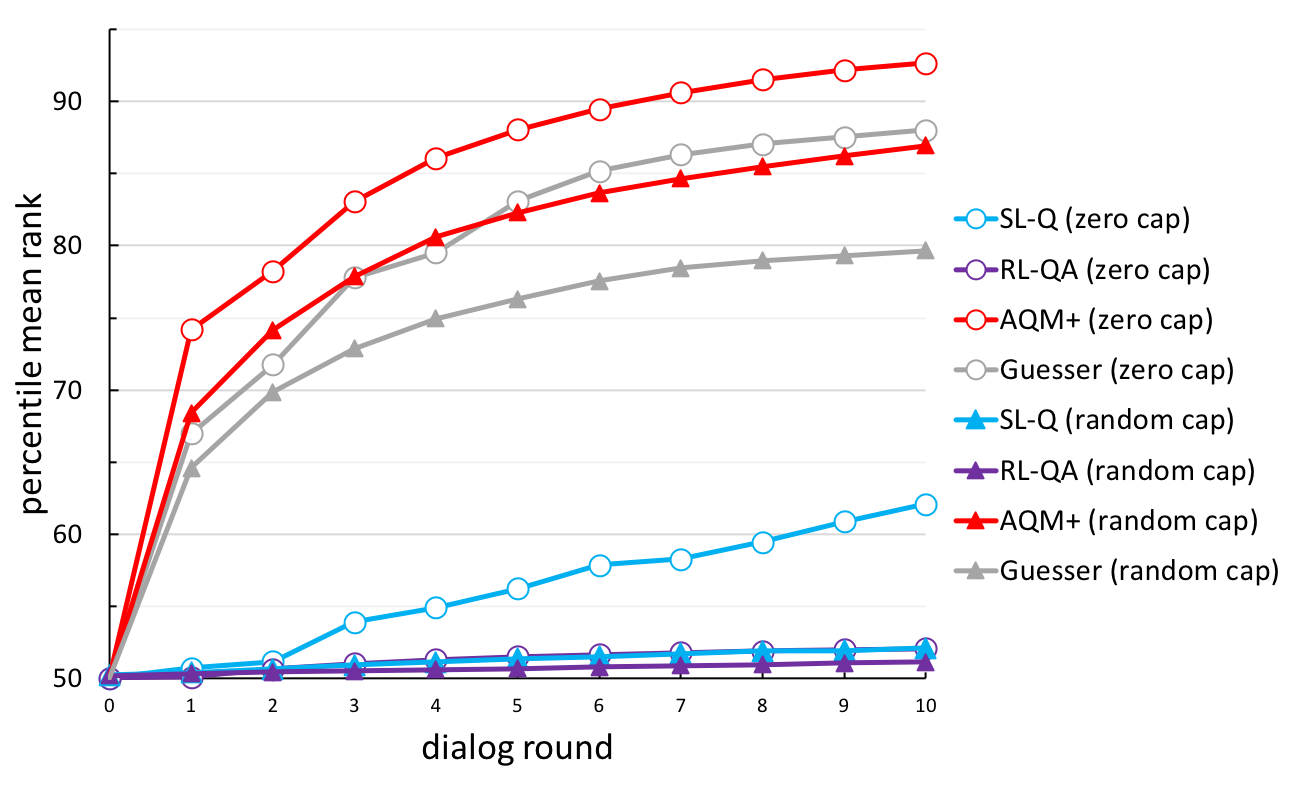}
\end{subfigure}
\begin{subfigure}[b]{0.48\textwidth}
\caption{Random Candidate Answers (Non-delta)}
\includegraphics[width=\textwidth]{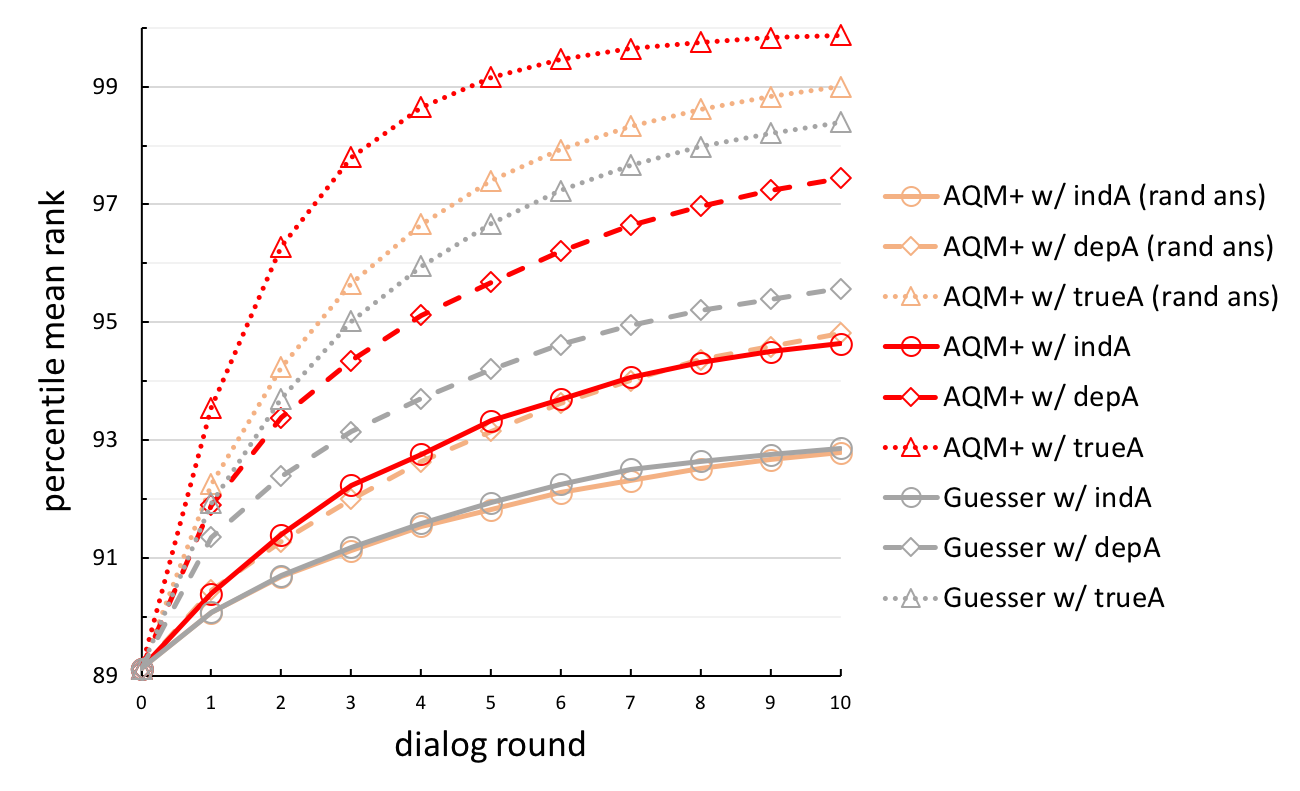}
\end{subfigure}
\caption{Left column shows the results of ablation studies on no caption experiments. Right column shows the result of ablation studies on random candidate answers experiments, where candidate answers are sampled from the training data.}
\label{fig:iclr19_fig3_1}
\end{figure}

\begin{figure}[t] 
\centering
\begin{subfigure}[b]{0.48\textwidth}
\caption{Number of QAC Experiment (indA, Non-delta)}
\includegraphics[width=\textwidth]{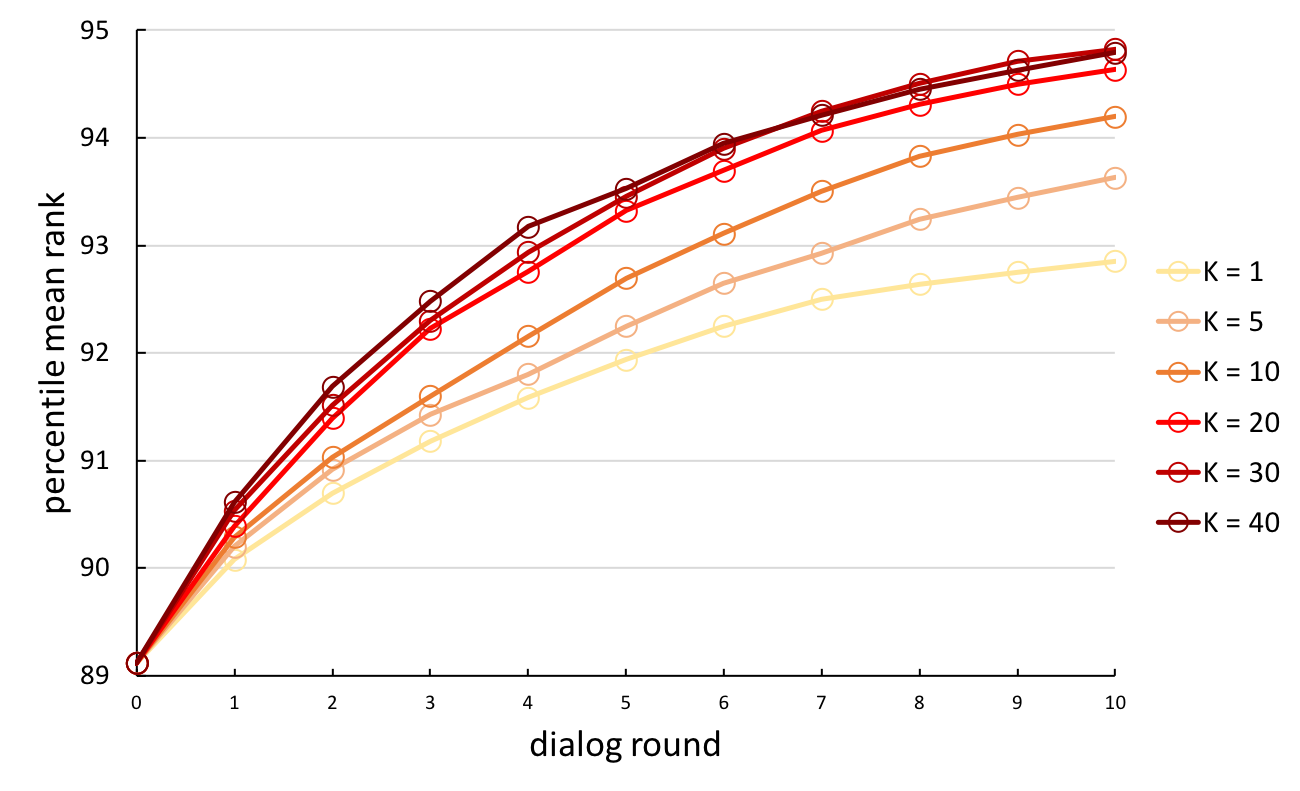}
\end{subfigure}
\begin{subfigure}[b]{0.48\textwidth}
\caption{Number of Q Experimnet (indA, Non-delta)}
\includegraphics[width=\textwidth]{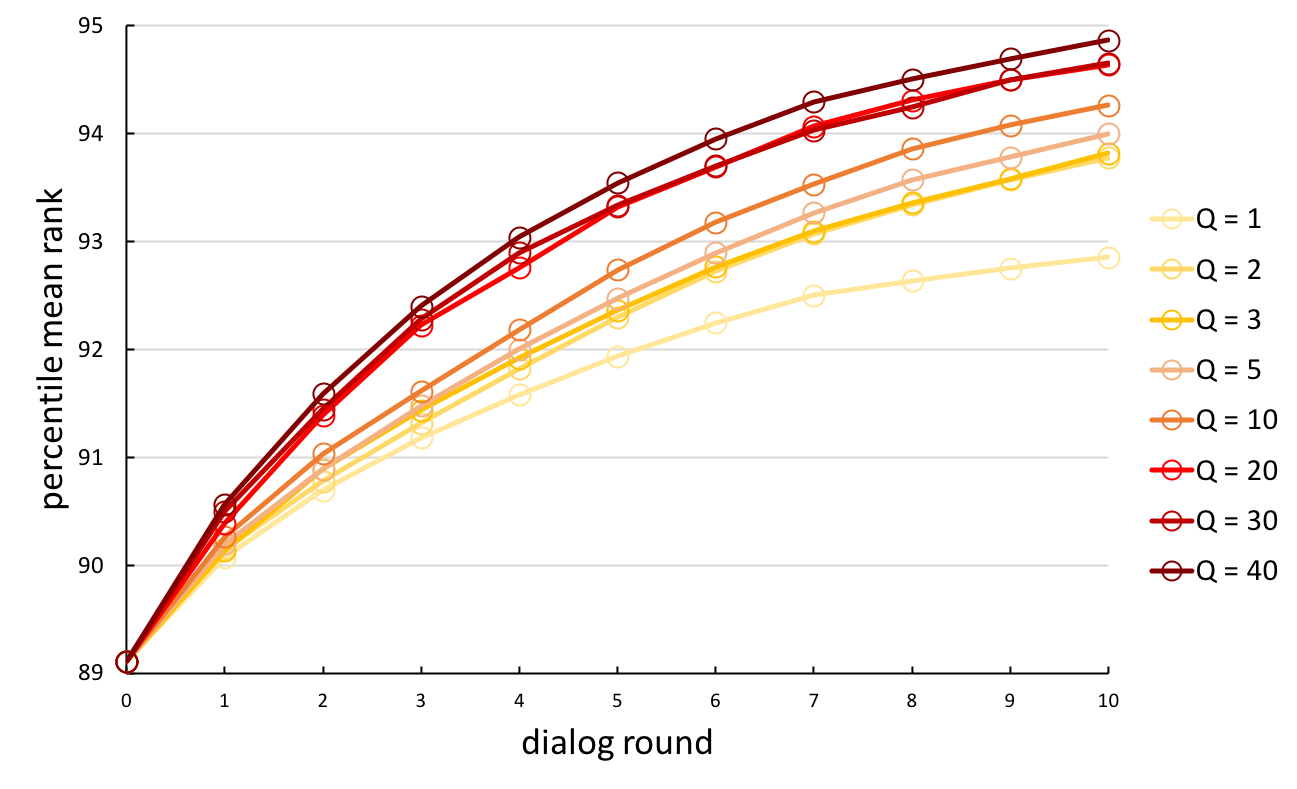}
\end{subfigure}
\begin{subfigure}[b]{0.48\textwidth}
\caption{Number of A Experiment (indA, Non-delta)}
\includegraphics[width=\textwidth]{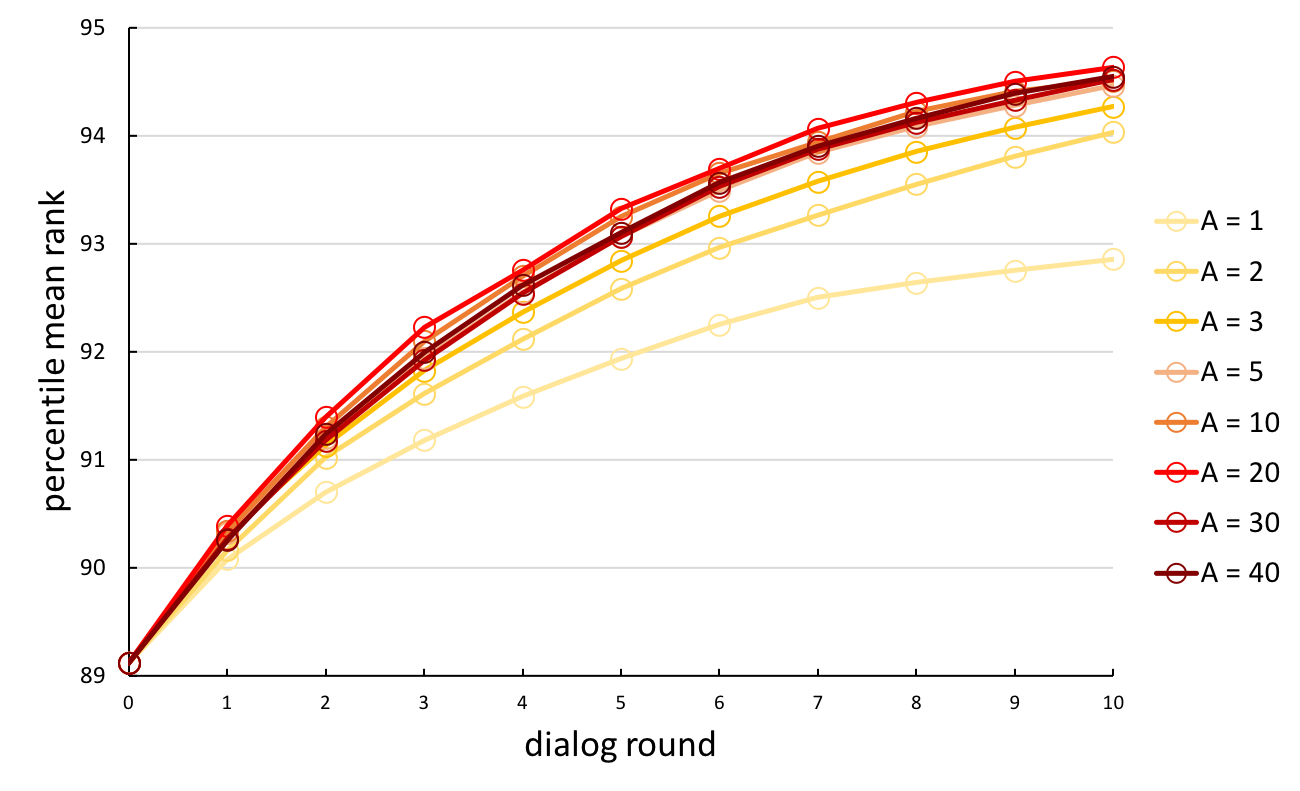}
\end{subfigure}
\begin{subfigure}[b]{0.48\textwidth}
\caption{Number of C Experimnet (indA, Non-delta)}
\includegraphics[width=\textwidth]{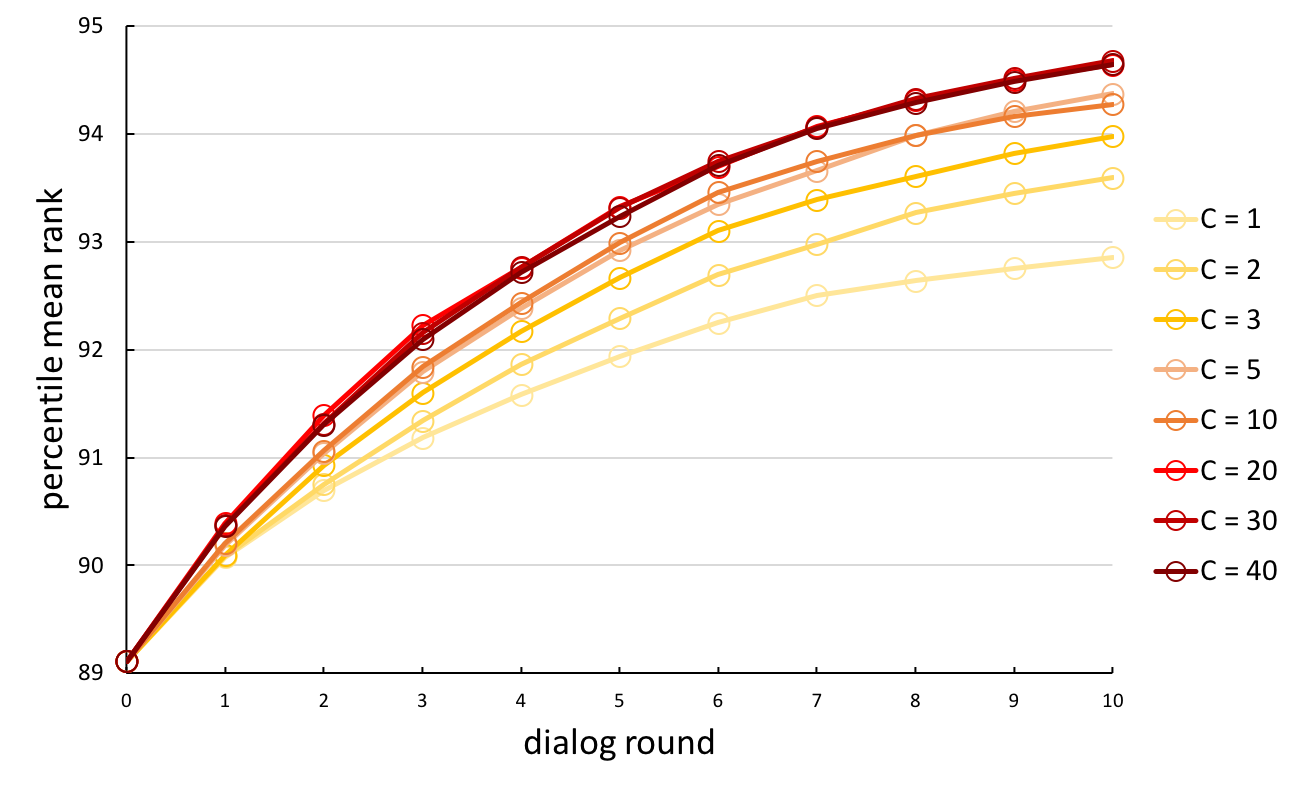}
\end{subfigure}
\caption{The result of ablation studies on different sizes of the subsets of candidate questions, answers, and classes. In the subfigure (a), the size for three subsets are the same to $K$.}
\label{fig:iclr19_fig3_2}
\end{figure}

\textbf{No Caption Experiment} 
We test our AQM+ algorithm where no caption information exists.
For the zeroth prediction, we simply replace the prior function from Qscore with a uniform function. Since Qgen in either SL-Q or RL-QA is trained also assuming the existence of the caption, we tried two alternative settings to approximate experiments without a caption. The first trial is the zero-caption experiment, where the caption vector is filled with zeros. 
The second trial is the random-caption experiment, where the caption vector is replaced with a random caption vector, which is not related to the target image. 
Figure \ref{fig:iclr19_fig3_1}a shows that AQM+ performs well for both zero-caption and random-caption setting.
By contrast, SL-Q and RL-QA do not work at all. It seems SL-Q and RL-QA are not trained on the situation where zero-caption vector or even totally wrong caption vector comes.
Though training SL-Q and RL-QA for these situations can increase their performance, it is evident that SL and RL algorithms are not robust to unexpected environments. 
Likewise, we also run no caption experiments for depA setting. For more ablation studies, see Figure \ref{fig:iclr19_fig6_} in Appendix B.

\textbf{Random Candidate Answers Experiment }
One of our main arguments is that generating candidate questions from Qscore and candidate answers from aprxAgen at every turn makes AQM+ effectively deal with general and complicated task-oriented dialogs.
Supporting the argument, we conducted the experiments under the setting where the answer set is randomly selected from the training data and then fixed. Random selection of candidate answers decreases the performance from 94.64\% to 92.78\% at indA, non-delta, and the 10th round.
Appendix B also includes a discussion on the setting with a predefined candidate question set $\mathbf{Q}_{fix}$.

\textbf{Number of QAC Experiment} 
We also changed the size of subset $K$ = $|\mathbf{Q}_{t,gen}|$=$|\mathbf{A}_{t,topk}(q_t)|$=$|\mathbf{C}_{t,topk}|$ to check our efficiency of information gain approximation, using non-delta setting.
Figure \ref{fig:iclr19_fig3_2}a shows the experimental results.
Note that AQM+ with the setting of $K=1$ corresponds to Guesser.
In the setting of non-delta and indA, 94.64\% of PMR is achieved when $K$ is 20, whereas 94.79\% is achieved when $K$ is 40. Note that 8 times (2 x 2 x 2) complexity increase just improves 0.15\% of PMR, showing the efficiency of the setting of $K$=20 in our experiments.
On the other hand, this result also implies that increasing $K$ would make further improvement on the performance.
Likewise, in depA setting, changing $K$ from 20 to 40 increases the PMR from 97.44\% to 97.77\%. For more ablation studies, see Figure \ref{fig:iclr19_fig7_} in Appendix B.
We also changed the size of each subset, $|\mathbf{Q}_{t,gen}|$, $|\mathbf{A}_{t,topk}(q_t)|$, and $|\mathbf{C}_{t,topk}|$. Figure \ref{fig:iclr19_fig3_2}b-d shows the results. $|\mathbf{Q}_{t,gen}|$ has the most effect, whereas $|\mathbf{A}_{t,topk}(q_t)|$ has the least effect.

\begin{figure}[t] 
\centering
\includegraphics[width=1.00\textwidth]{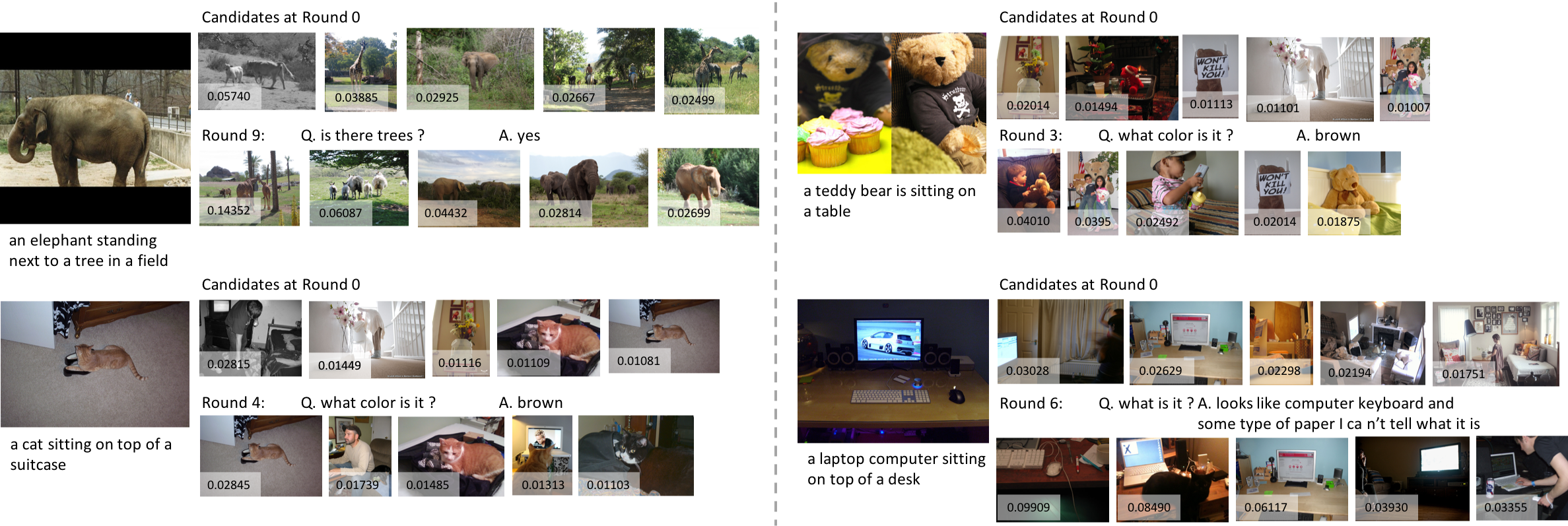}
\caption{Qualitative results on image retrieval of AQM+. Left column shows true images and their corresponding caption, and right column contains selected top-$k$ images.}
\label{fig:iclr19_fig4}
\end{figure}

\textbf{Generated Questions and Selected Images }
Figure \ref{fig:iclr19_fig4} shows the top-k images selected by AQM+'s posterior. Non-delta and indA setting is used.  
The figure shows that relevant images to the caption remained after few dialog turns. The bottom number in the image denotes posterior of the image AQM+ thinks of.
We also compare selected examples of generated dialog of SL-Q, RL-QA, and AQM+ w/ indA for delta setting. See Figure \ref{fig:iclr19_fig8_} in Appendix C for the results.

\section{Discussion}

\subsection{Difficulty of GuessWhich }
According to our results, we infer that PMR degradation of comparative SL and RL models during the dialog is not caused by forgetting dialog context to ask an appropriate question.
Comparative results between AQM+ and Guesser show that the improvement from AQM+'s Qpost is significant, which implies that the major constraint of SL and RL is the limited capacity of RNN and its softmax score function.

Another reason for the poor performance lies in the current status of VQA models. According to \cite{das2017a}, 
they discovered a variety of models, one of which is used in both the study of \cite{das2017b} and our experiments, and
can already reach 41.2\% for answer retrieval accuracy from 100 candidate answers, solely using the question without exploiting image and history information. Fully exploiting these factors, however, increases the performance only slightly to 45.5\%. 
As discrimination on different images relies on image and history information, Qbot suffers to gain meaningful information through the dialog.
Therefore, applying AQM+ to the GuessWhich problem means that we not only solve a very complicated problem, but also find that the AQM framework is applicable to the situation where the answer has high uncertainty.

\subsection{Notes on Comparative Analysis}

\textbf{Fine-tuning both Qbot and Abot through RL }
Though RL-QA is the main setting in the work of \cite{das2017b}, there are some reports indicating that fine-tuning both Qbot and Abot is unfair \citep{de2017,han2017}\rtt{, as one of the ultimate goals in this field is to make a questioner be able to talk with human.}
If the distribution of Abot is not fixed during RL, Qbot and Abot can make their own language which is not compatible to natural language \citep{kottur2017}.
To prevent this problem, many studies added the objective function of language model during RL \citep{zhu2017,das2017b}.
However, even though the generated dialog is tuned to be like human dialog, the performance of RL-QA on the conversation with human would decrease compared to SL-Q, because the distribution of Abot become far from human's \citep{chattopadhyay2017,lee2018}.
Moreover, achieving a good performance by fine-tuning both Qbot and Abot is much easier than fine-tuning only Qbot \citep{zhu2017,han2017}.
Thus, it is reasonable to compare AQM+ w/ indA and AQM+ w/ depA with SL-Q and RL-Q, respectively.

\textbf{Compuational Cost} 
AQM+ at $K$=20 uses 20$\times$20$\times$20 calculations for information gain. On the other hand, the previous AQM requires 20$\times \infty \times$9628 calculations for information gain, which makes the computation intractable. Even if we use only 100 candidate answers, which is in the Visual Dialog dataset \citep{das2017a}, the previous AQM requires 2500 times as many calculations (20M) as AQM+.
On the other hand, AQM+ requires more calculations and thus requires more inference time than SL or RL. AQM+ generates one question within around 3s when $K$=20, whereas SL generates one question within 0.1s. 
We used Tesla P40 for our experiments. Though the complexity of our information gain is $O(K^3)$, $K$ does not increase the time required for the whole inference in proportion to the cube of K, when $K$=20. 
It is because calculating the information gain is not the sole resource-intensive part in the whole inference process.

\subsection{Toward Practical Applications }
There are plenty of potential future works to improve the performance of AQM+ in real task-oriented dialog applications.
For example, robust task-oriented dialog systems are required for appropriately replying to user's questions \citep{li2017} and responding for chit-chat style conversation \citep{zhao2017}. The question quality can also be improved by diverse beam search approaches \citep{vijayakumar2016,li2016}, which prevent sampling similar questions for the candidate set. We highlight two issues described below; online learning and fast inference.

\textbf{Online Learning} For a novel answerer, fine-tuning on the dialog model is required \citep{Krause2018}. If the experiences of many users are available, model-agnostic meta learning (MAML) \citep{finn2017} can be applied for few-shot learning. Updating the hyperparameter $\lambda$ in an online manner, which balances the effect of the prior and the likelihood, can also be effective in practice. If the answer distribution of user is different from our aprxAgen, we can increase $\lambda$ to decrease the effect of the likelihood.

\textbf{Fast Inference} AQM+'s time complexity can be decreased further by changing the structure of aprxAgen. In specific, we can apply diverse methods such as skipping the update of hidden states in some steps \citep{seo2018}, using convolution networks or self-attention networks \citep{yu2018b,vaswani2017}, substituting matrix multiplication operation for hidden state update to weighted addition \citep{yu2018a}, and direct information gain inference from the neural networks \citep{belghazi2018}.

\section{Conclusion}
Asking appropriate questions in practical applications has recently been paid attention \citep{rao2018,buck2018}.
We proposed AQM+ algorithm that is a large-scale extension of AQM framework. AQM+ can ask an appropriate question considering the context of the dialog, handle the responses in a sentence form, and efficiently estimate information gain of the target class with a given question. This improvement makes our AQM framework to step forward toward practical task-oriented applications.
AQM+ not only outperforms the comparative SL and RL algorithms, but also enlarges the gap between AQM+ and the comparative algorithms comparing to the performance gaps reported in GuessWhat.
AQM+ acheives more than 60\% error decreases through the dialog, whereas the comparative algorithms only achieve 6\% error decreases.
Moreover, the performance of AQM+ can be boosted further by employing the models recently proposed in the visual dialog field such as other question generator models \citep{jain2018} and question answering models \citep{kottur2018}.

\subsubsection*{Acknowledgments}
\rtt{The authors would like to thank Yu-Jung Heo, Hwiyeol Jo, and Kyunghyun Cho for helpful comments.
This work was supported by the Creative Industrial Technology Development Program (10053249) funded by the Ministry of Trade, Industry and Energy (MOTIE, Korea).}

\small
\setlength{\bibsep}{5.0pt}
\bibliography{iclr2019_conference}
\bibliographystyle{iclr2019_conference}
\normalsize

\newpage

\section*{Appendix A. AQM+ Algorithm}

The question generating process of AQM+ used in our GuessWhich experiments are as follows.

\begin{algorithm}[h]
   \caption{Question Generating Process of AQM+ in Our GuessWhich Experiments}
   \label{alg:alg1}
\begin{algorithmic}
    \STATE $\hat{p}'(c|h_0)$ $\propto$ $\exp(\lambda \cdot f^\ddagger(c|h_0))$
    \FOR {$t$ = 1:$T$}
        \STATE $\mathbf{C}_{t,topk}$ $\leftarrow$ top-K posterior test image (from Qpost $\hat{p}(c|h_{t-1})$)
        \STATE $\mathbf{Q}_{t,gen}$ $\leftarrow$ top-K likelihood questions using beam search (from Qgen $p^\dagger(q_t|h_{t-1})$)
        \STATE $\mathbf{A}_{t,topk}(q_t)$ $\leftarrow$ generated answers from aprxAgen for question $q_t$ and each class in $\mathbf{C}_{t,topk}$ (from aprxAgen $\tilde{p}(a_t|c,q_t,h_{t-1})$)
        \STATE $q_t \leftarrow $ argmax$_{q'_t \in \mathbf{Q}_{t,gen}}$ $\tilde{I}[C,A_t;q'_t,a_{1:t-1},q_{1:t-1}]$ with $\mathbf{A}_{t,topk}(q_t)$ and $\mathbf{C}_{t,topk}$ in Eq. \ref{eq:eq1}
        \STATE Get $a_t$ from Agen $\bar{p}(a_t|c,q_t,h_{t-1})$ 
        \STATE Update Qpost $\hat{p}(c|h_t) \propto \tilde{p}(a_t|c,q_t,h_{t-1}) \cdot \hat{p}(c|h_{t-1})$ in Eq. \ref{eq:eq2}
    \ENDFOR
\end{algorithmic}
\end{algorithm}

\section*{Appendix B. Ablation Study}

\begin{figure}[h] 
\centering
\begin{subfigure}[b]{0.48\textwidth}
\caption{Number of QAC Experiment (depA, Non-delta)}
\includegraphics[width=\textwidth]{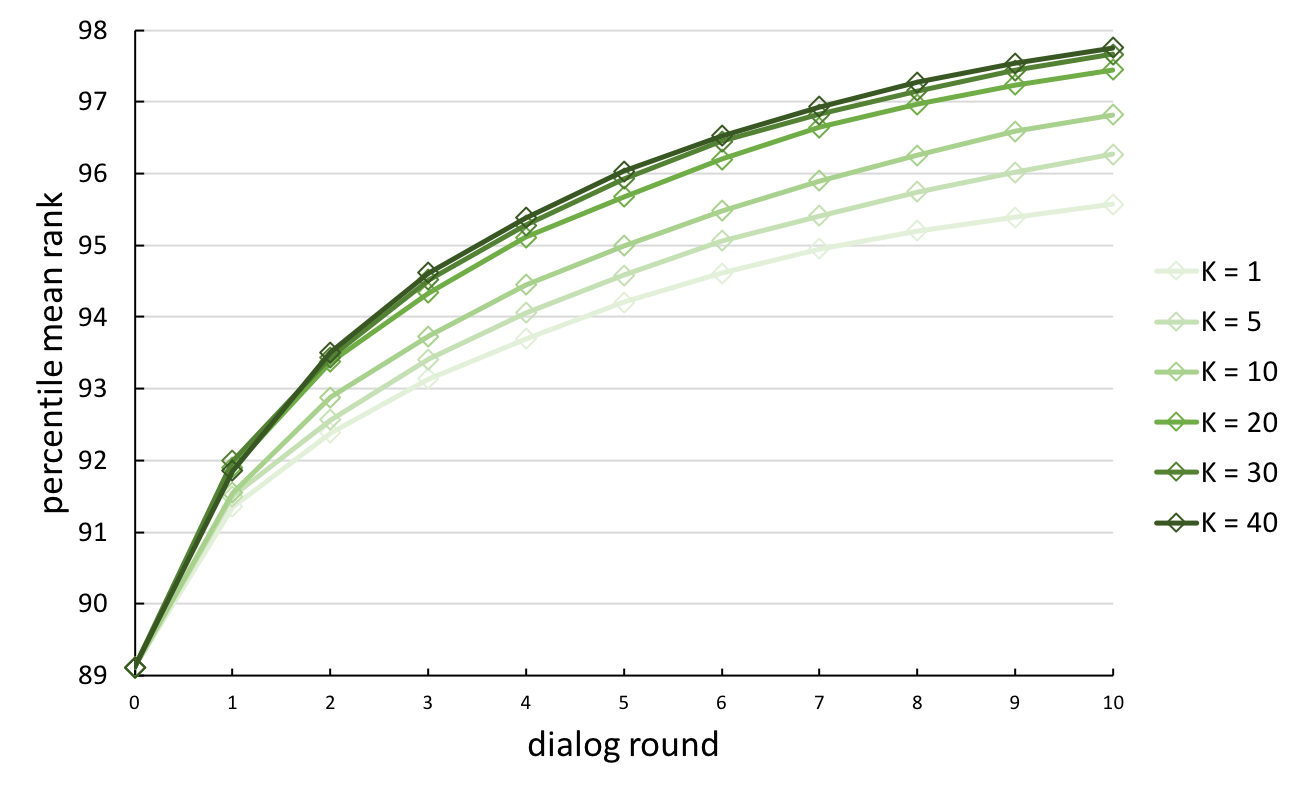}
\end{subfigure}
\begin{subfigure}[b]{0.48\textwidth}
\caption{Number of QAC Experiment (trueA, Non-delta)}
\includegraphics[width=\textwidth]{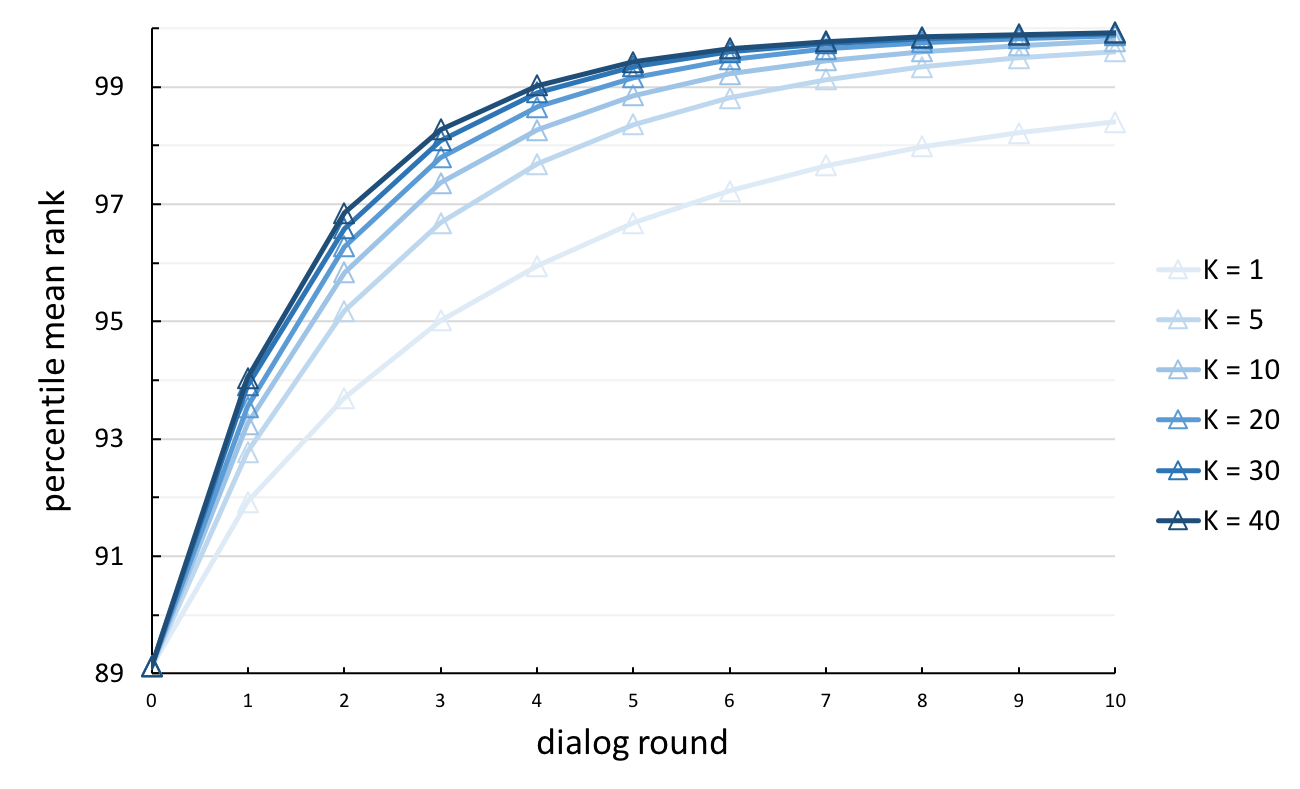}
\end{subfigure}
\caption{Ablation study on different sizes of the subset of candidate questions, answers, and classes. The size for three subsets are the same to K. The results of the non-delta setting with depA and trueA are illustrated.}
\label{fig:iclr19_fig6_}
\end{figure}

\begin{figure}[h] 
\centering
\begin{subfigure}[b]{0.48\textwidth}
\caption{No Caption Experiment (depA, Non-delta)}
\includegraphics[width=\textwidth]{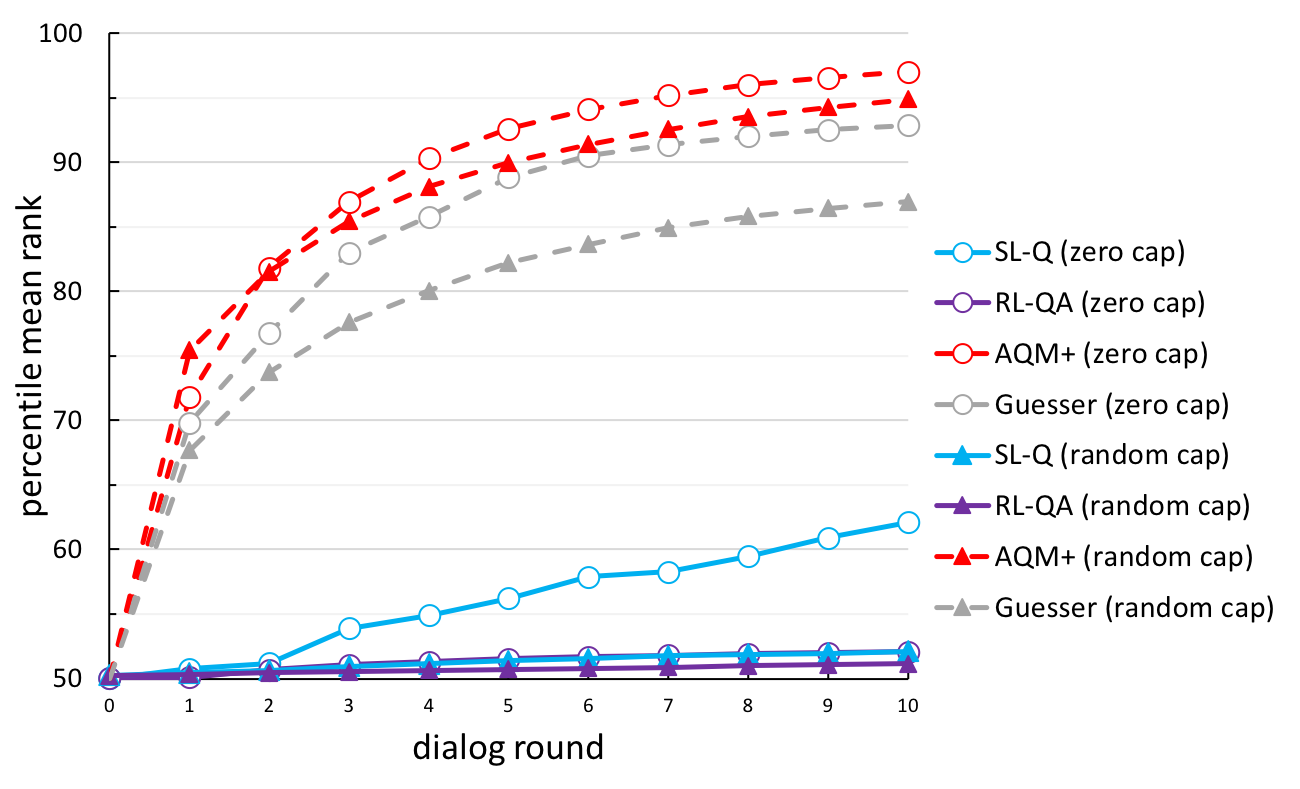}
\end{subfigure}
\begin{subfigure}[b]{0.48\textwidth}
\caption{No Caption Experiment (trueA, Non-delta)}
\includegraphics[width=\textwidth]{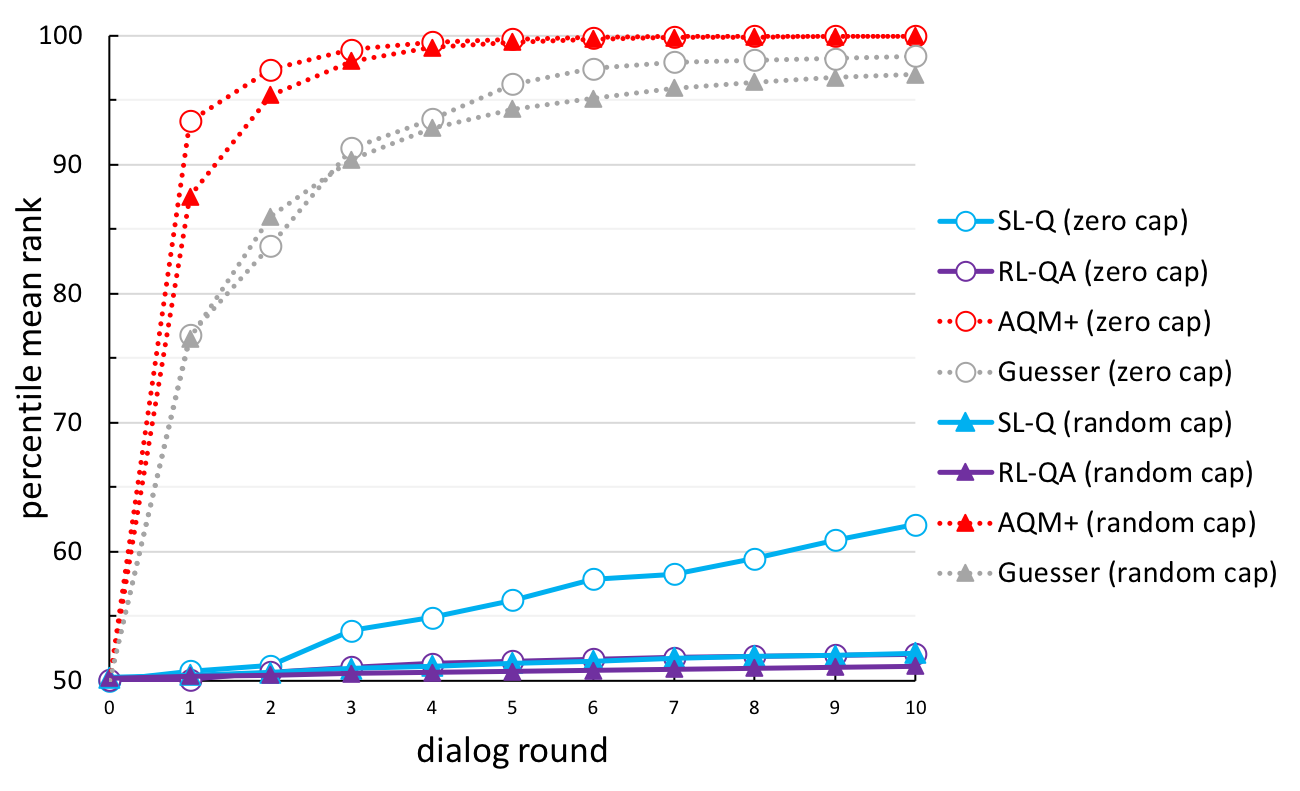}
\end{subfigure}
\caption{Ablation study on no caption experiment. The results of the non-delta setting with depA and trueA are illustrated.}
\label{fig:iclr19_fig7_}
\end{figure}

Figure \ref{fig:iclr19_fig6_} shows the results of the number of QAC ablation experiment on depA and trueA, in the non-delta setting. The effect of K decreases in trueA compared to indA, which indicates that the similarity between the distribution of aprxAgen and Agen is related to the effectiveness of large K.
Figure \ref{fig:iclr19_fig7_} shows the results of the no caption experiment on depA and trueA, in the non-delta setting.

\begin{figure}[h] 
\centering
\begin{subfigure}[b]{0.48\textwidth}
\caption{AQM+'s Qinfo + SL's Qscore (Non-delta)}
\includegraphics[width=\textwidth]{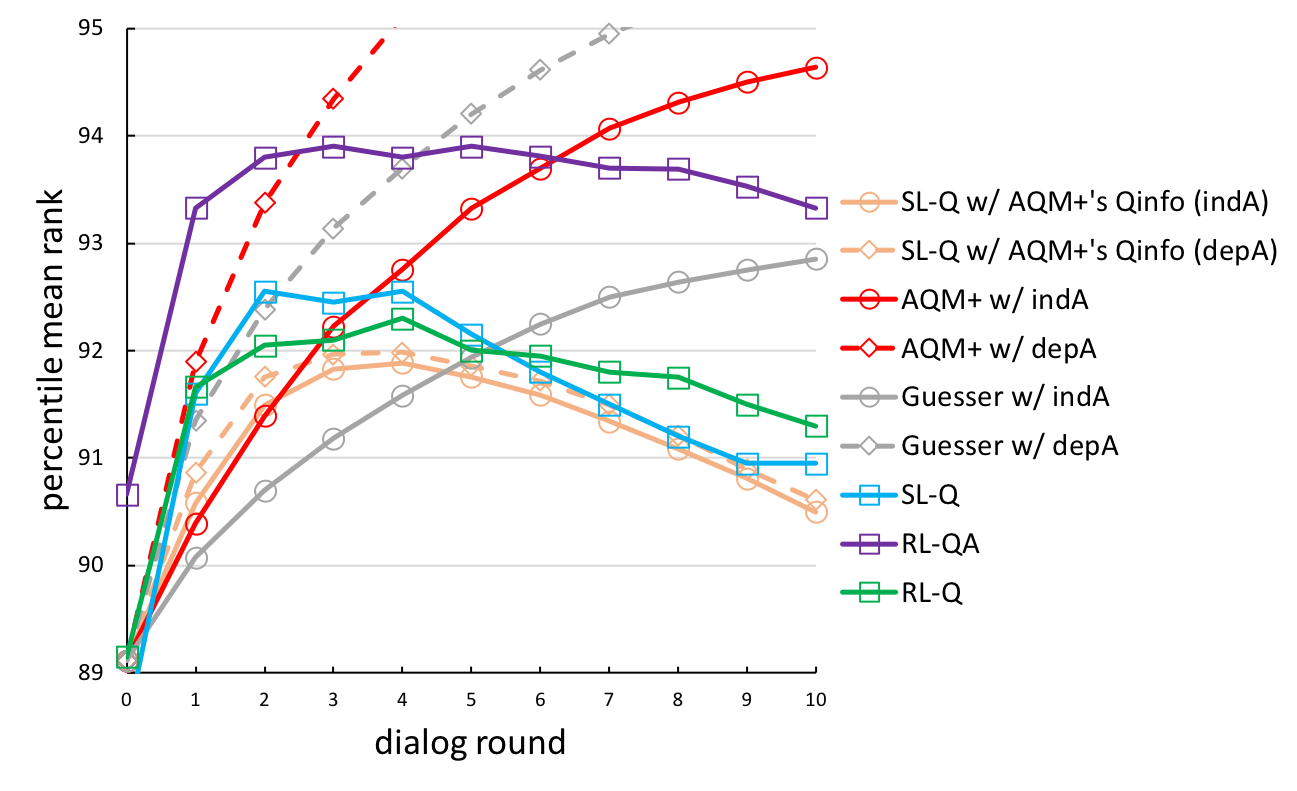}
\end{subfigure}
\begin{subfigure}[b]{0.48\textwidth}
\caption{AQM+'s Qinfo + SL's Qscore (Delta)}
\includegraphics[width=\textwidth]{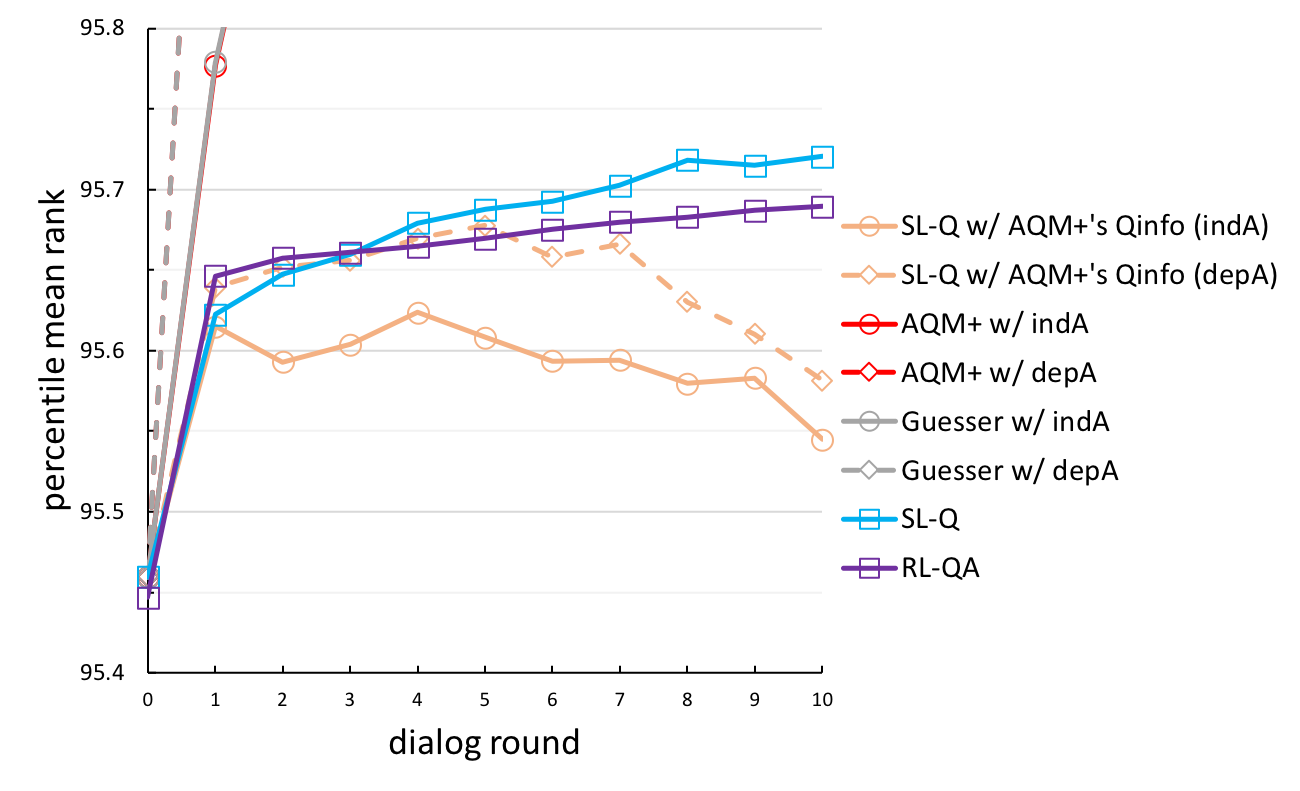}
\end{subfigure}
\caption{Ablation study on the model with AQM+'s question-generator and SL's guesser.}
\label{fig:iclr19_fig11_}
\end{figure}

Figure \ref{fig:iclr19_fig11_} shows the experimental results on the model where AQM+'s Qinfo is used as the question-generator and SL's Qscore is used as the guesser.
AQM+’s Qinfo does not improve the performance of SL’s guesser (Qscore). 
Our analysis of the results is as follows.
For delta setting, the SL guesser is not able to obtain the information from the answers. 
For the non-delta case, not dialog history but caption information gives dominant information to SL’s guesser. The questions which often appear with caption thus gave a more clear signal for the target class for SL’s guesser. Figure \ref{fig:iclr19_fig11_}a shows that SL-Q performs better than RL-Q in the early phase, but SL-Q’s performance decreases faster than that of RL-Q in the later phase. It is because SL-Q generates the question to be more likely to have co-appeared with the caption than RL-Q. Likewise, AQM+’s question does not help SL’s guesser because AQM+ generates questions that are more independent of the caption.

\begin{figure}[h] 
\centering
\begin{subfigure}[b]{0.48\textwidth}
\caption{gen1Q Experiment (Non-delta)}
\includegraphics[width=\textwidth]{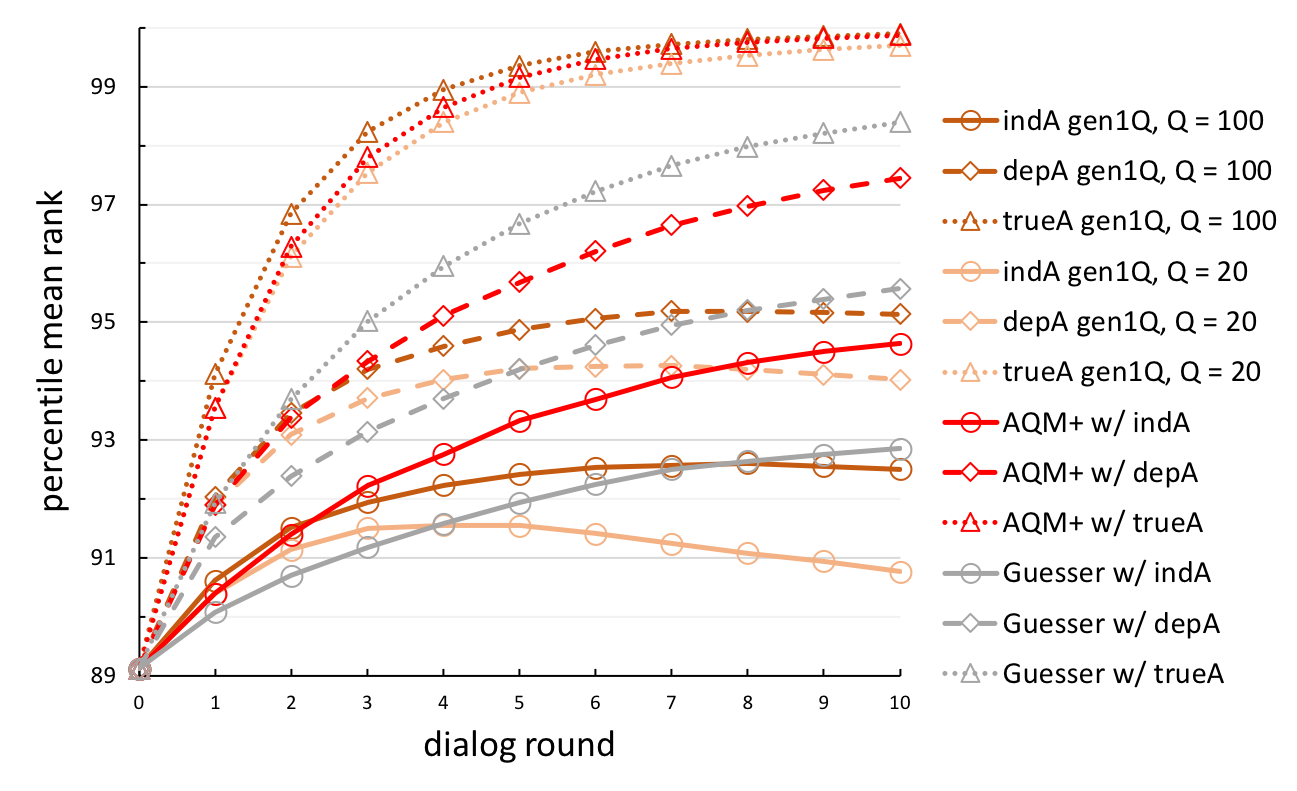}
\end{subfigure}
\begin{subfigure}[b]{0.48\textwidth}
\caption{gen1Q Experiment (Delta)}
\includegraphics[width=\textwidth]{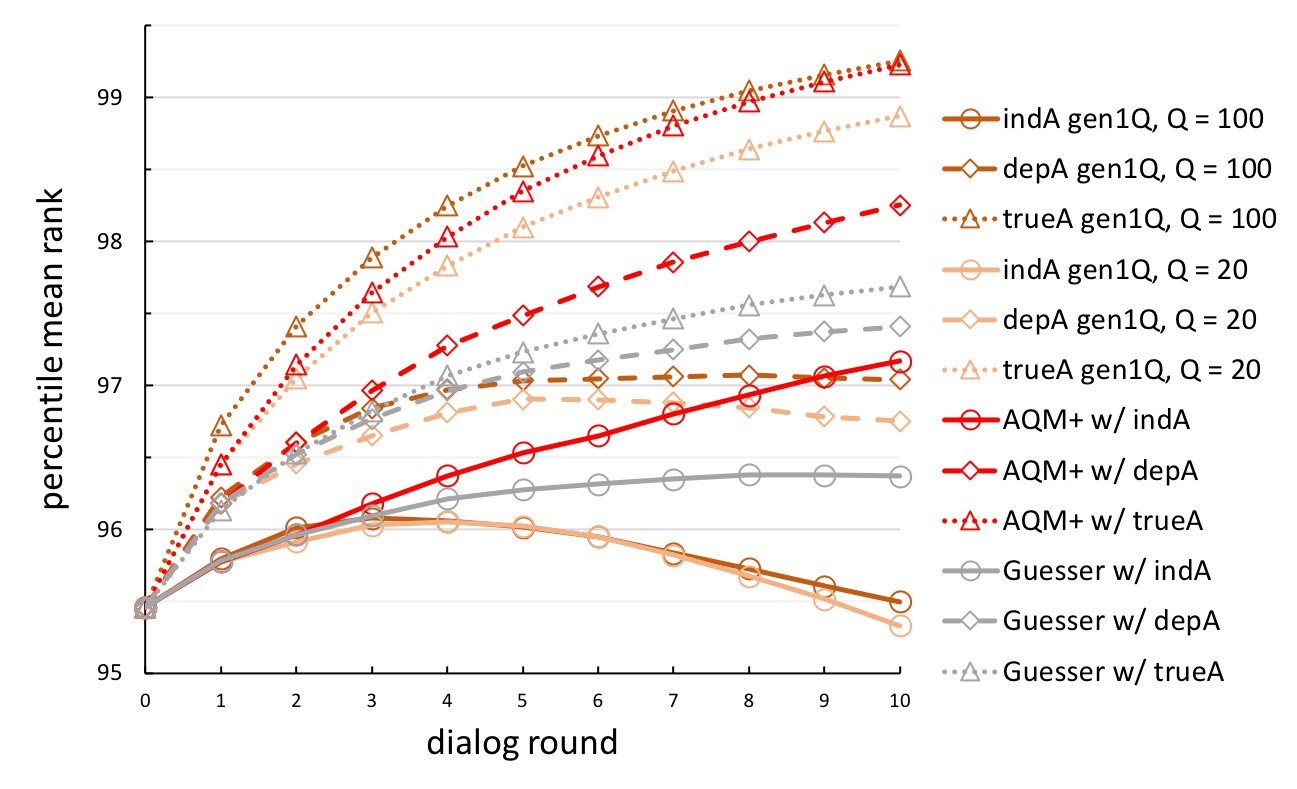}
\end{subfigure}
\caption{Ablation study on gen1Q. The candidate questions are generated only at the first turn.}
\label{fig:iclr19_fig8_}
\end{figure}

\begin{figure}[h] 
\centering
\begin{subfigure}[b]{0.48\textwidth}
\caption{randQ Experiment (Non-delta)}
\includegraphics[width=\textwidth]{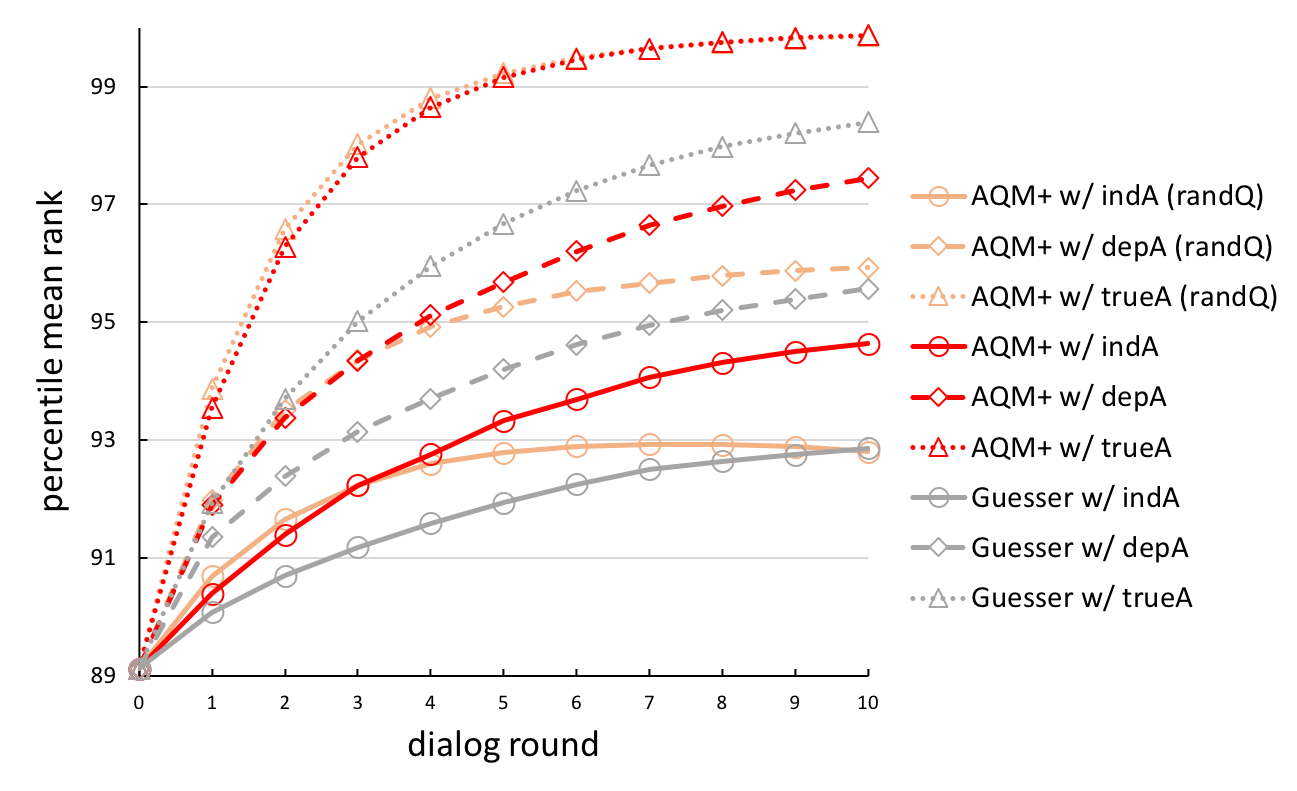}
\end{subfigure}
\begin{subfigure}[b]{0.48\textwidth}
\caption{randQ Experiment (Delta)}
\includegraphics[width=\textwidth]{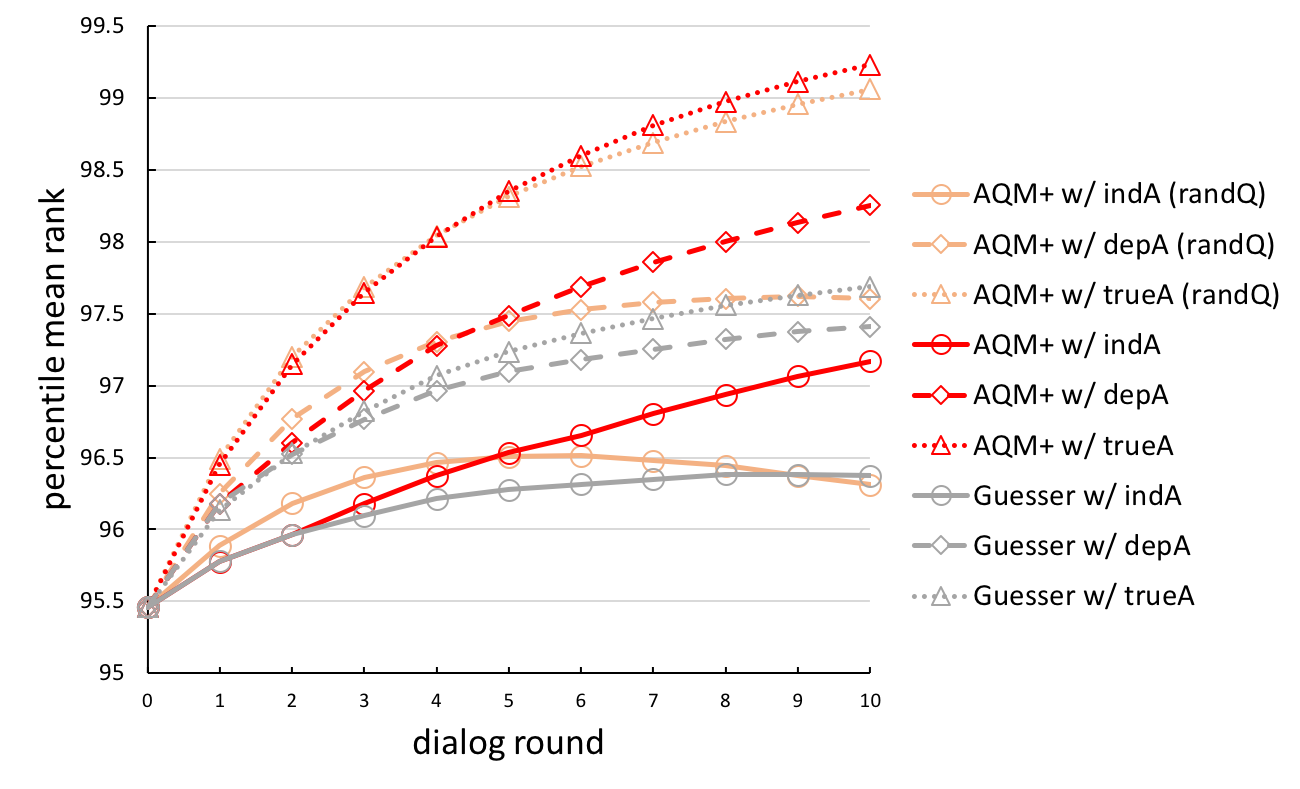}
\end{subfigure}
\caption{Ablation study on randQ. The candidate questions are extracted from the training data.}
\label{fig:iclr19_fig9_}
\end{figure}

We conducted the experiments under the setting where a predefined candidate question set $\mathbf{Q}_{fix}$ is used. The discussion section in the work of \cite{lee2018} includes an experimental setting in which the candidate questions are generated from an end-to-end SL model only at the first turn. We refer to this setting as gen1Q, as in the previous AQM paper. Figure \ref{fig:iclr19_fig8_} shows the results of gen1Q ablation study. Note that this setting of $|\mathbf{Q}|$=100 requires five times as many computations to calculate the information gain as the original AQM+, despite gen1Q performs even worse than Guesser baseline. 
Another noticeable phenomenon is that there is no significant performance loss in trueA setting. Since aprxAgen in trueA knows the exact probability of Abot's answer, by exploiting such an aprxAgen, Qbot in trueA can clearly distinguish between different classes by capturing even the subtle differences in answer distributions given similar questions.
We also performed the experiments under the setting where $\mathbf{Q}_{fix}$ comes from training data. Figure \ref{fig:iclr19_fig9_} shows the results of randQ ablation study. The baseline method with this $\mathbf{Q}_{fix}$ showed accuracy degradation.
Regardless of the PMR, we point out that randQ retrieves questions relevant to neither the caption nor the target image. It is why we generate candidate questions from a seq-to-seq model.

\begin{figure}[h] 
\centering
\begin{subfigure}[b]{0.48\textwidth}
\caption{No History Experiment (Non-delta)}
\includegraphics[width=\textwidth]{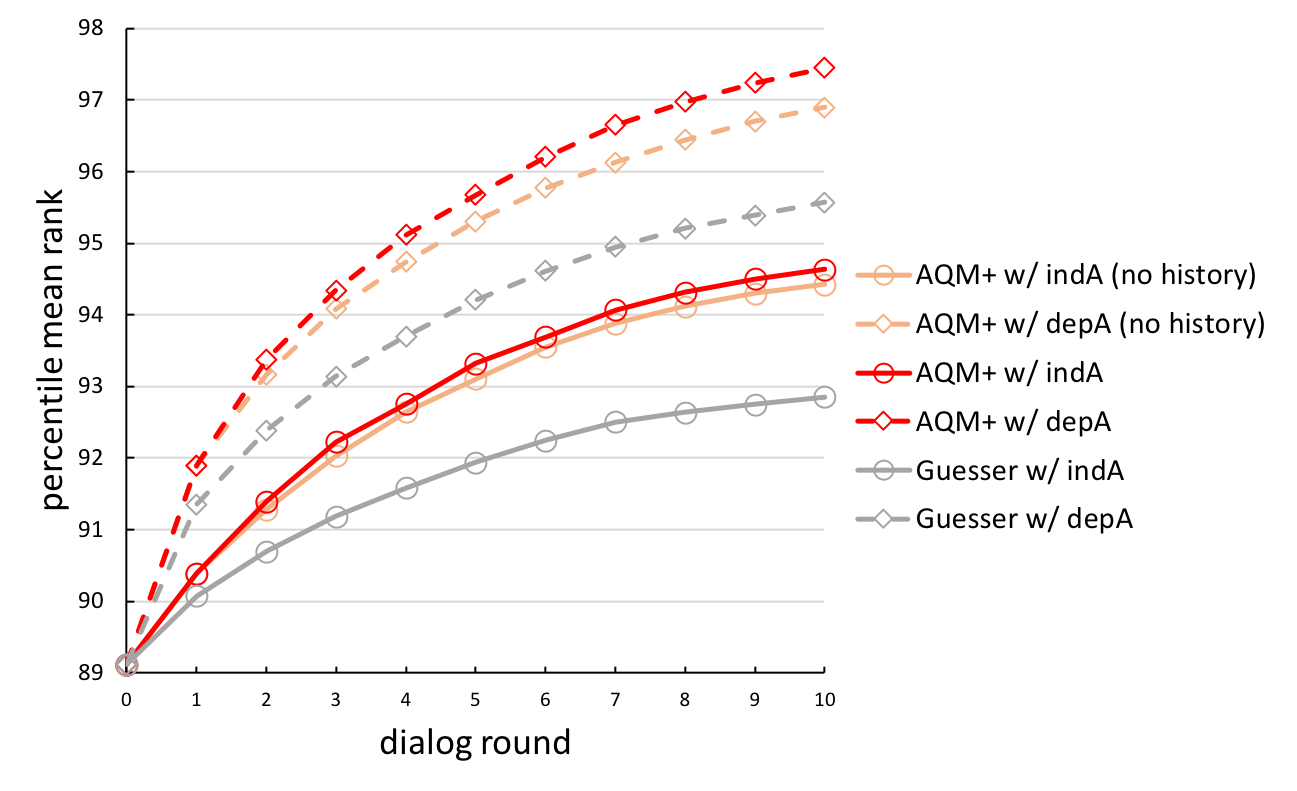}
\end{subfigure}
\begin{subfigure}[b]{0.48\textwidth}
\caption{No History Experiment (Delta)}
\includegraphics[width=\textwidth]{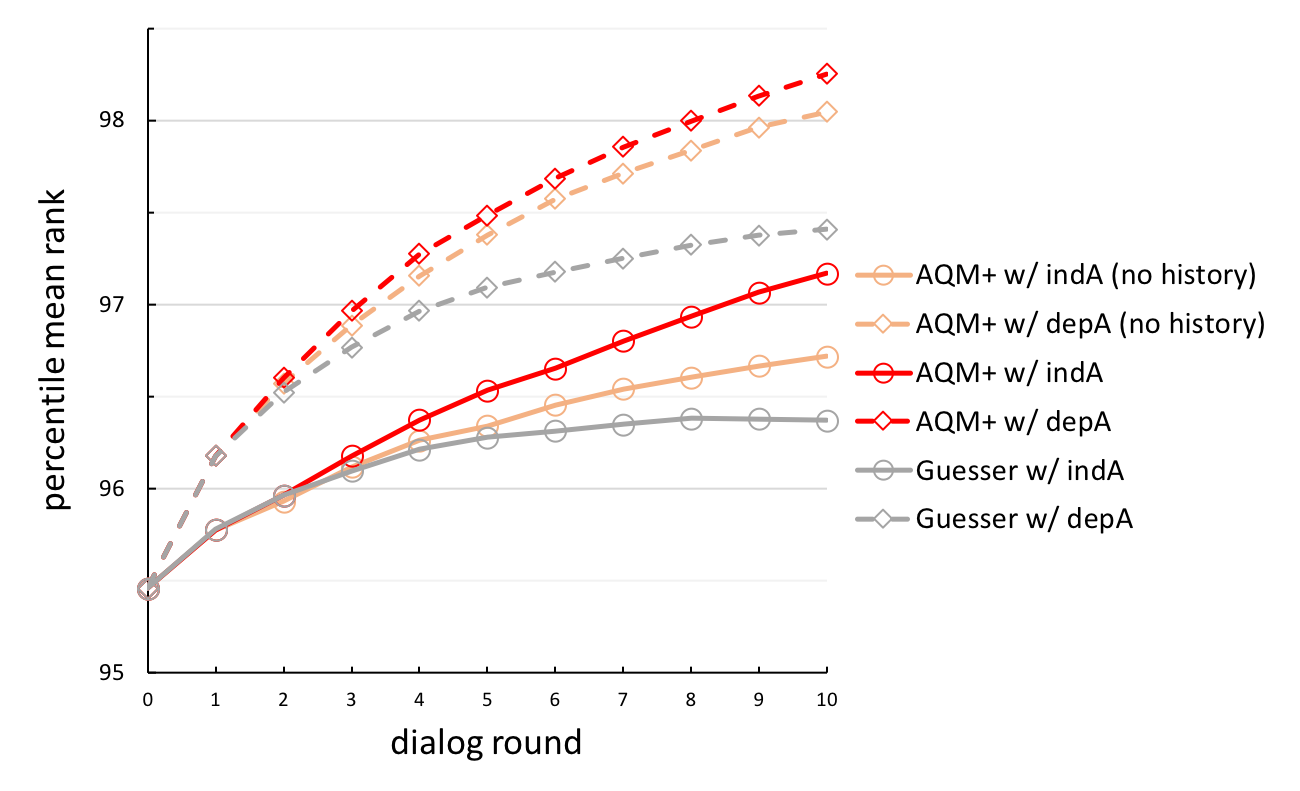}
\end{subfigure}
\caption{Ablation study on no history experiment. Under this setting, aprxAgen ignores the dialog history.}
\label{fig:iclr19_fig10_}
\end{figure}

Figure \ref{fig:iclr19_fig10_} shows the results of the no history experiment. Dialog history helps to guess the target image but is not critical. Ablating history makes the performance decrease by 0.22\% and 0.56\% for indA and depA in non-delta, respectively, and 0.46\% and 0.21\% for indA and depA in delta, respectively.

\newpage
\textcolor[rgb]{1,1,1}{.}

\newpage
\textcolor[rgb]{1,1,1}{.}


\newpage

\section*{Appendix C. Generating Sentences}

\begin{figure}[h] 
\centering
\includegraphics[width=1.00\textwidth]{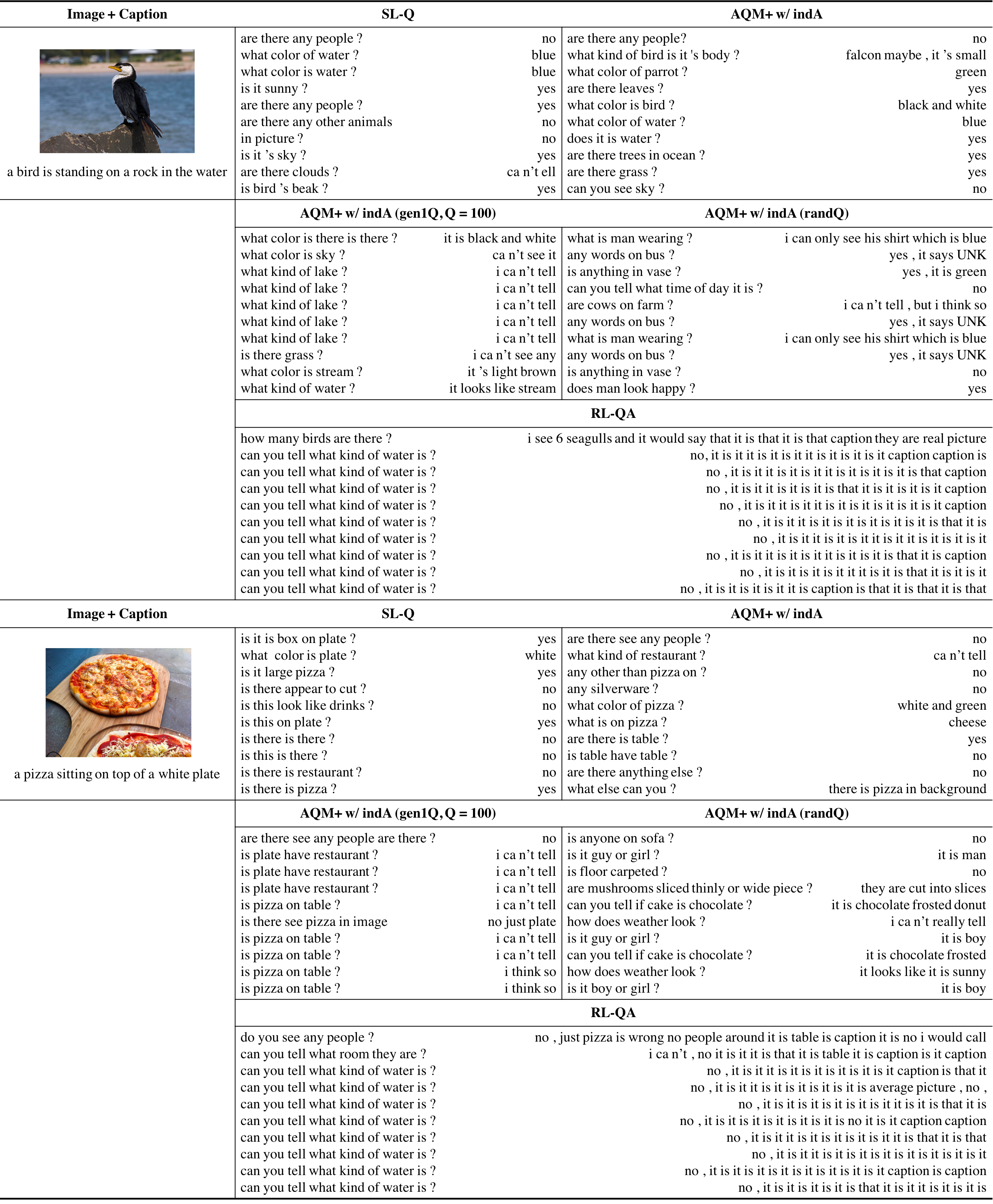}
\caption{Selected examples of generated dialog in delta setting.}
\label{fig:iclr19_fig12_}
\end{figure}

Figure \ref{fig:iclr19_fig12_} shows selected examples of generated questions in delta setting. 
Though delta setting boosts to increase PMR of the zeroth turn much, it degenerates the question quality, especially for RL-QA.
Moreover, RL-QA tends to concentrate on the first turn, leaving questions and answers of the remaining turns meaningless.

\end{document}